\documentclass[journal]{vgtc}                     

\onlineid{0}

\vgtccategory{Research}

\vgtcpapertype{please specify}

\title{BondMatcher: H-Bond Stability
Analysis
in Molecular Systems}

\author{%
  Thomas Daniel,
  Malgorzata Olejniczak, and
  Julien Tierny
}

\authorfooter{
  \item
    Thomas Daniel and Julien Tierny are with the CNRS and Sorbonne University.
  	E-mail: \{firstname.lastname\}@sorbonne-universite.fr
  \item
    Malgorzata Olejniczak is with the Centre of New Technologies, the University of Warsaw.
    E-mail: malgorzata.olejniczak@cent.uw.edu.pl
}

\abstract{This is an abstract, line1.\\
line 2.\\
line 3.\\
line 4.\\
line 5.\\
line 6.\\
line 7.\\
line 8.\\
line 9.\\
line 10.\\
line 11.\\
line 12.}

\abstract{This application paper investigates the stability of hydrogen
bonds 
\julien{(H-bonds)},
as characterized by the Quantum Theory of Atoms in Molecules (QTAIM).
First, we contribute a database of 4544 electron densities
associated to four
\gosia{isomers of} water hexamers (the so-called \emph{Ring},
\emph{Book}, \emph{Cage} and \emph{Prism}), generated by distorting their
equilibrium geometry under various \gosia{structural}
perturbations, modeling the
natural dynamic behavior of molecular systems.
Second, we present a new stability measure, called \emph{bond occurrence
rate}, associating each
\julien{bond path}
present
at
equilibrium
with its rate of
occurrence within the input ensemble.
We also provide an algorithm, called \emph{BondMatcher}, for its automatic
computation, based on a tailored, geometry-aware partial isomorphism estimation
between
the extremum graphs of the considered
electron densities.
Our new stability measure allows for the automatic identification of
densities lacking
\julien{H-bond}
paths,
enabling further visual inspections. Specifically,
the topological
analysis
enabled by our framework
corroborates experimental observations
and provides refined geometrical criteria for characterizing the disappearance
of
\julien{H-bond paths}.
Our electron density database and our C++ implementation are available
at this address:
\href{https://github.com/thom-dani/BondMatcher}{https://github.com/thom-dani/BondMatcher}.
}


\teaser{
  \vspace{-0.5ex}
  \centering
  \includegraphics[width=\linewidth]{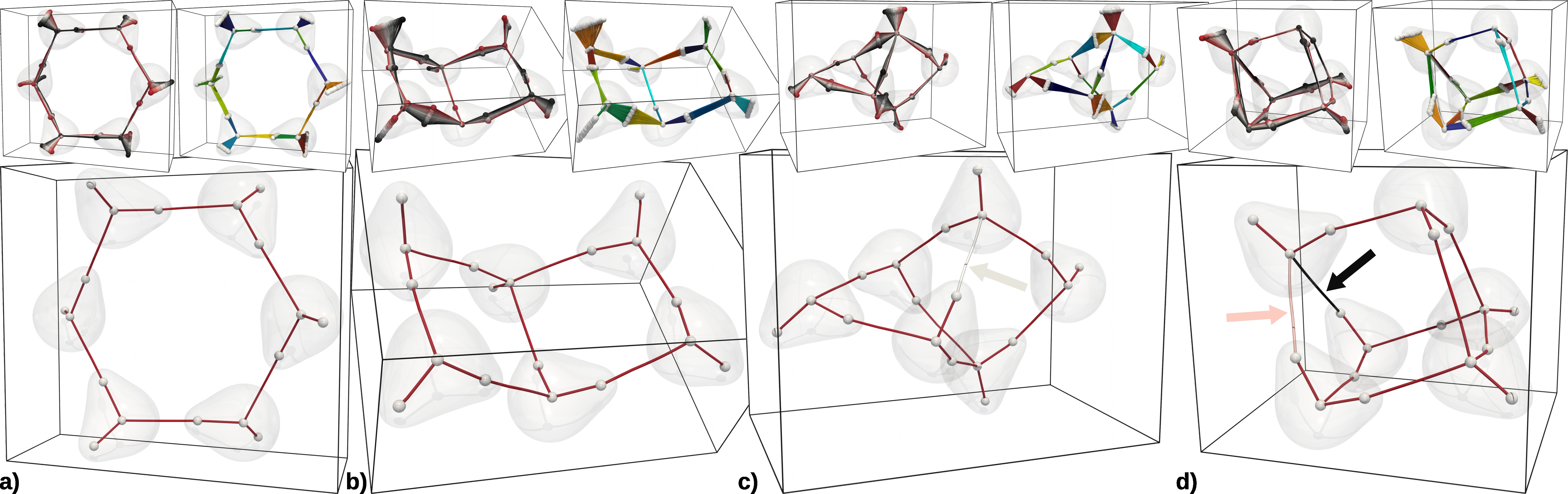}
  \caption{H-bond stability analysis under vibrational displacements for
  $4$
isomers of the water hexamer: \emph{(a)} the \emph{Ring} (vibration
mode $0$),
\emph{(b)}
the \emph{Book} (vibration
mode $0$),
\emph{(c)}
the \emph{Cage} (vibration
mode $18$), and
\emph{(d)}
the \emph{Prism} (vibration
mode $0$). The
$6$
water molecules are shown with transparent isosurfaces.
\emph{Top:}
Electron density extremum graphs for the input
displacements (left, black to red), partial isomorphisms to the
equilibrium
(right, colored arcs, white: unmatched arcs).
\emph{Bottom:} Equilibrium extremum graphs,
colored by bond occurrence rate (black  to white  to red). Our
analysis identifies H-bonds
paths
(arrows) susceptible of breaking under
molecular
vibrations (e.g., turning the \emph{Cage} into the \emph{Book}, white
bond). Our per-bond matching
enables a detailed analysis of bonding
indicators through the ensemble, corroborating experimental observations and
refining geometrical criteria
for the disappearance of H-bonds.}
  \label{fig:teaser}
}

\graphicspath{{figs/}{figures/}{pictures/}{images/}{./}} 

\usepackage{tabu}                      
\usepackage{booktabs}                  
\usepackage{lipsum}                    
\usepackage{mwe}                       

\usepackage{mathptmx}                  
\usepackage{amsmath}
\usepackage{amssymb}

\DeclareMathOperator*{\argmin}{arg\,min}

\DeclareMathAlphabet{\pazocal}{OMS}{zplm}{m}{n}
\SetMathAlphabet\pazocal{bold}{OMS}{zplm}{bx}{n}

\begin{document}



\renewcommand{\mathcal}[1]{\pazocal{#1}}

\newcommand{\electronDensity}{\rho}

\newcommand{\surface}{S}

\newcommand{\extremumGraph}{\mathcal{E}}
\newcommand{\minimumSet}{\mathcal{N}}
\newcommand{\unstableSets}{\mathcal{A}}
\newcommand{\occurrence}{\mathcal{O}}

\newcommand{\signal}{\mathcal{S}}
\newcommand{\noise}{\mathcal{N}}
\newcommand{\simplex}{\sigma}
\newcommand{\domain}{\mathcal{K}}
\newcommand{\numberOfVertices}{n_v}
\newcommand{\numberOfSimplices}{n_\simplex}
\newcommand{\dataVector}{v}
\newcommand{\dataVectorSpace}{\mathcal{V}}
\newcommand{\filtration}{\mathcal{F}}
\newcommand{\persistenceMap}{\mathcal{P}}
\newcommand{\energy}{\mathcal{E}}
\newcommand{\loss}{\mathcal{L}}
\newcommand{\range}{\mathbb{R}}
\newcommand{\sublevelset}[1]{#1^{-1}_{-\infty}}
\newcommand{\superlevelset}[1]{#1^{-1}_{+\infty}}
\newcommand{\Star}{St}
\newcommand{\Link}{Lk}
\newcommand{\diagram}{\mathcal{D}}
\newcommand{\target}{\diagram_T}
\newcommand{\complexity}{\mathcal{O}}

\newcommand{\face}{\tau}
\newcommand{\lowerlink}{\Link^{-}}
\newcommand{\upperlink}{\Link^{+}}
\newcommand{\Index}{\mathcal{I}}
\newcommand{\offset}{o}
\newcommand{\Natural}{\mathbb{N}}
\newcommand{\criticalSet}{\mathcal{C}}

\newcommand{\wasserstein}[1]{\mathcal{W}_{#1}}
\newcommand{\projection}{\Delta}
\newcommand{\hierarchy}{\mathcal{H}}
\newcommand{\decimation}{D}
\newcommand{\xDimD}{L_x^\decimation}
\newcommand{\yDimD}{L_y^\decimation}
\newcommand{\zDimD}{L_z^\decimation}
\newcommand{\xDim}{L_x}
\newcommand{\yDim}{L_y}
\newcommand{\zDim}{L_z}
\newcommand{\cubicalComplex}{\mathcal{C}}
\newcommand{\Grid}{\mathcal{G}}
\newcommand{\GridD}{\mathcal{G}^\decimation}
\newcommand{\x}{\phantom{x}}
\newcommand{\Mod}{\;\mathrm{mod}\;}
\newcommand{\NN}{\mathbb{N}}
\newcommand{\forwardIntegralLine}{\mathcal{L}^+}
\newcommand{\backwardIntegralLine}{\mathcal{L}^-}
\newcommand{\triangulationOp}{\phi}
\newcommand{\decimationOp}{\Pi}
\newcommand{\isovalue}{w}
\newcommand{\persistence}{p}
\newcommand{\pointMetric}{d}
\newcommand{\diagramSet}{\mathcal{S}_\mathcal{D}}
\newcommand{\diagramSpace}{\mathbb{D}}
\newcommand{\jointree}{\mathcal{T}^-}
\newcommand{\splittree}{\mathcal{T}^+}
\newcommand{\mergetree}{\mathcal{T}}
\newcommand{\tree}{\mergetree}
\newcommand{\depth}{d}
\newcommand{\mergetreeSet}{\mathcal{S}_\mathcal{T}}
\newcommand{\branchset}{\mathcal{S}_\mathcal{B}}
\newcommand{\branchspace}{\mathbb{B}}
\newcommand{\mergetreeSpace}{\mathbb{T}}
\newcommand{\editdistance}{D_E}
\newcommand{\wassersteinTree}{W^{\mergetree}_2}
\newcommand{\distanceSequence}{d_S}
\newcommand{\branchtree}{\mathcal{B}}
\newcommand{\branchtreeSet}{\mathcal{S}_\mathcal{B}}
\newcommand{\branchtreeSpace}{\mathbb{B}}
\newcommand{\forest}{\mathcal{F}}
\newcommand{\sequenceSpace}{\mathbb{S}}
\newcommand{\forestMatrix}{\mathbb{F}}
\newcommand{\treeMatrix}{\mathbb{T}}
\newcommand{\normalizedLocation}{\mathcal{N}}
\newcommand{\normalizedWasserstein}{W^{\normalizedLocation}_2}
\newcommand{\geodesictree}{\mathcal{G}}
\newcommand{\dummyVector}{\mathcal{V}}
\newcommand{\geodesictreeVec}{g}
\newcommand{\geodesicAxis}{\mathcal{A}}
\newcommand{\directionVector}{\mathcal{V}}
\newcommand{\geodesicdiagram}{\mathcal{G}^{\diagram}}
\newcommand{\reconstructionError}{E_{L_2}}
\newcommand{\pcaBasis}{B_{\mathbb{R}^d}}
\renewcommand{\pcaBasis}{B}
\newcommand{\origin}{o_b}
\newcommand{\sizeEncoding}{n_e}
\newcommand{\sizeDecoding}{n_d}
\newcommand{\linearTransformation}{\psi}
\newcommand{\unitTransformation}{\Psi}
\renewcommand{\origin}{o}
\newcommand{\bdtOrigin}{\mathcal{O}}
\newcommand{\activation}{\sigma}
\newcommand{\validBDT}{\gamma}
\newcommand{\mtPgaBasis}{B_{\branchtreeSpace}}
\newcommand{\mtPgaError}{E_{\wassersteinTree}}
\newcommand{\frechetEnergy}{E_F}
\newcommand{\geodesicExtremity}{\mathcal{E}}
\newcommand{\vectorNotation}[1]{\protect\vv{#1}}
\renewcommand{\vectorNotation}[1]{#1}
\newcommand{\axisNotation}[1]{\protect\overleftrightarrow{#1}}
\newcommand{\individualEnergy}{E}
\newcommand{\ensembleSize}{N}
\newcommand{\numberBranchinBarycenter}{N_1}
\newcommand{\numberGeodesicSamples}{N_2}
\newcommand{\planarGridX}{N_x}
\newcommand{\planarGridY}{N_y}
\newcommand{\regularGrid}{G}
\newcommand{\distanceMatrix}{\mathbb{D}}
\newcommand{\maxDimensions}{{d_{max}}}
\newcommand{\projectionOperator}{\mathcal{P}}
\newcommand{\reconstructed}[1]{\widehat{#1}}
\newcommand{\gt}{>}
\newcommand{\lt}{<}
\newcommand{\branch}{b}
\newcommand{\nonLinearFunction}{\sigma}
\newcommand{\batchSequence}{S}
\newcommand{\homologyGroup}{\mathcal{H}}
\newcommand{\bettiNumber}{\beta}
\newcommand{\still}{\mathcal{S}}

\newcommand{\julien}[1]{\textcolor{blue}{#1}}
\renewcommand{\julien}[1]{\textcolor{black}{#1}}

\newcommand{\gosia}[1]{\textcolor{purple}{#1}}
\renewcommand{\gosia}[1]{\textcolor{black}{#1}}

\newcommand{\revision}[1]{\textcolor{blue}{#1}}
\renewcommand{\revision}[1]{\textcolor{black}{#1}}
\newcommand{\minor}[1]{\textcolor{blue}{#1}}

\newcommand{\discuss}[1]{\textcolor{black}{#1}}

\renewcommand{\figureautorefname}{Fig.}
\renewcommand{\sectionautorefname}{Sec.}
\renewcommand{\subsectionautorefname}{Sec.}
\renewcommand{\equationautorefname}{Eq.}
\renewcommand{\tableautorefname}{Tab.}
\newcommand{\algorithmautorefname}{Alg.}
\newcommand{\lineautorefname}{Alg.}

\newcommand{\todo}[1]{\textcolor{red}{#1}}
\renewcommand{\todo}[1]{\textcolor{black}{#1}}

\newcommand{\mycaption}[1]{
\caption{#1}
}

\newcommand{\myparagraph}[1]{
\noindent\textbf{#1}}

\newcommand{\journal}[1]{\textcolor{blue}{#1}}
\renewcommand{\journal}[1]{\textcolor{black}{#1}}

\newcommand{\myspace}{\vspace{-0.0775ex}}

\newcommand{\myspaceFour}{\vspace{-0.35ex}}




\firstsection{Introduction}

\maketitle

In chemistry, the properties of a molecular system are determined by the
interactions between the atoms forming the system.
Two main types of interactions are typically distinguished.
First,
\emph{covalent bonds},
refer to interactions that arise from the sharing of electrons between atoms
\cite{zhao.etal_nrc_2019a}.
Second,
\emph{non-covalent interactions} denote all other types of chemical
interactions.
These interactions are more subtle but  they
%
play a crucial role in many biological processes
and
chemical design
tasks (e.g., molecular folding, protein docking). Among the
documented non-covalent bonds
\cite{taylor_cg&d_2024},
hydrogen
bonds (or \emph{H-bonds})
\cite{arunan.etal_pac_2011}
form a prominent bond type which intervenes in
many
molecular systems, from simple water dimers to macromolecules
such
as DNA.

\revision{Nevertheless,}
H-bonds are challenging to study as they are weaker
than covalent bonds. As a consequence, they can easily
rearrange under molecular motions
such as vibrations or rotations. Then, the condition of appearance and
disappearance of H-bonds
under such structural rearrangements is an active research topic,
both theoretically and experimentally \cite{gao.etal_jcs_2024}. 
Such a flexibility challenges the structural description of H-bonds,
which often relies
on simple, distance-based, heuristic criteria
for establishing their presence,
without
explicit consideration of the underlying electronic effects
\cite{arunan.etal_pac_2011}.

In contrast, the
\emph{Quantum Theory of Atoms In Molecules} (QTAIM) \cite{bader94, matta07}
provides a framework to describe molecular interactions at
the electronic level thanks to the analysis of the electron
density field.
QTAIM,
combined with robust computational tools
based on discrete Morse theory (DMT) \cite{forman98, harshChemistry, ttk17,
ttk19},
enables a quantitative, reliable
and interpretable analysis of chemical interactions
\cite{chemistry_vis14, Malgorzata19}. Still, to the best of our knowledge,
these robust computational tools based on DMT
have only been used for the \emph{static} analysis of molecular
systems, in particular at their equilibrium state.
However, considering the natural \emph{dynamic} behavior of molecular systems
(e.g., vibrations, rotations) is of paramount importance, in particular to
assess the stability of H-bonds.

In this work, we fill this gap 
and extend the topological techniques based on DMT and QTAIM
\cite{harshChemistry, ttk17, ttk19}
to characterize 
the dynamic
behavior of molecular systems. 
For this, we introduce a new framework, called \emph{BondMatcher},
which robustly and automatically matches bonds across multiple dynamic states of
a given molecular system.
At a technical level, this approach exploits the
specific
hypotheses
about the molecular system
to reliably compute
relevant partial isomorphisms between the
\emph{extremum graphs}
of the electron density fields
(\autoref{sec_ensemble}). This automatic correspondence enables a
fine, quantitative and interpretable
analysis of the presence
of H-bond paths within molecular systems under various dynamical effects.
Specifically, we document the analysis of two
case studies
focusing on four
prominent, low-energy
\julien{isomers}
(namely, the \emph{Ring}, the
\emph{Book}, the \emph{Cage} and the \emph{Prism}) of water hexamer, a
three-dimensional system playing a central role for the understanding of the
different states of water.
The first
case study
(\autoref{sec_useCase_pathways})
focuses on
two rotational motions
documented in 
experiments \cite{rotations16}, each yielding an ensemble of $256$ scalar
fields modeling the electron density of the corresponding configurations. The
second
case study
(\autoref{sec_useCase_vibrations})
investigates
the $48$ intrinsic vibration modes of the above
isomers,
yielding $192$ ensembles of $21$ scalar fields each. In both cases,
our
analysis based on the topological description of the electron
density corroborates documented observations \cite{vibrations1, vibrations2,
rotations16}. Moreover, the fine scale of our approach (at the electronic level)
enables a qualitative interpretation of the geometric conditions favoring the
appareance of H-bond paths, resulting in refined
suggestions
for the geometrical criteria for H-bond detection in molecular graphs.
We
provide
as additional material
our C++ implementation,
along with
our
electron density database ($4544$ datasets, about $800$ gigabytes),
which may constitute a benchmark for future methods
dealing with
ensemble chemistry data.

\ifdefined\includeSuggestions
\color{purple}

\color{black}
\fi

\subsection{Related work}
\label{sec_relatedWork}

This section reviews the literature related 
to
our work, which can be
classified into the following main categories.

\myparagraph{Computational Quantum Chemistry:}
Quantum chemistry (QC) explores the properties of molecular systems
at the granularity of the interactions between electrons and nuclei.
Density functional theory (DFT) \cite{hohenberg64} is a prominent quantum
mechanical framework, based on a probabilistic representation of the
distribution of electrons in a molecular system, called the \emph{electron
density} (noted $\rho$). Based on this, Bader
introduced the \emph{Quantum Theory of Atoms in Molecules} (QTAIM)
\cite{bader94, matta07}, which
establishes a relation between the
critical points of the electron density  and various
chemical concepts (\autoref{sec_background_qtaim}).
QTAIM
plays a central role in modern
quantum chemistry as it enables a fine-scale interpretation and
understanding of many experimental observations.



\myparagraph{Molecular data analysis and visualization:}
In recent years, the visualization community has shown a substantial interest
in molecular data (as assessed by several surveys \cite{KozlikovaKFLBBV17}), as
its complexity presents unique challenges for rendering,
exploration and analysis. Several popular software platforms have emerged
\cite{avogadro, SkanbergHYL23} for the visualization and analysis of molecular
systems using ball-and-stick representations. 
In particular,
dedicated rendering techniques have been proposed for the interactive display
of these graphs for large-scale molecular systems \cite{chemistry12}.
Moreover, for dynamic contexts, specialized techniques have also been
documented for the tracking of consistent internal frames for these graphs
\cite{SkanbergLFHY19, SkanbergFLYH22}.
For even larger systems, molecular surfaces
\cite{chemistry15}
constitute an alternate
representation, based on an implicit surface derived from the molecule
specification, which facilitates the visualization and analysis of large-scale
phenomena (e.g., protein docking \cite{chemistry10,chemistry42},
cavities \cite{chemistry33, chemistry46}, pockets \cite{chemistry34,
chemistry58} and bonding \cite{chemistry37}). Specialized techniques for their
efficient rendering have also been documented
\cite{chemistry45, chemistry54}.
While they
provide a useful approximation of the shape of molecular systems,
molecular surfaces only model their geometry, without accounting for
the quantum mechanisms governing their structure. For a fine-scale analysis
of these mechanisms, several
techniques have been established on
top of specific scalar descriptors (e.g., the electron density). For instance,
approaches based on continuous scatterplots \cite{SharmaMTLHN24} or
Voronoi segmentations \cite{MasoodTLANH21} have been proposed
\revision{for charge transfer analysis.}

%
%

\myparagraph{Topological Data Analysis:}
Topological methods \cite{edelsbrunner09, zomorodianBook} have received a
considerable attention by the visualization community over
the last decades, given their
ability to robustly extract structural patterns within scalar data
\cite{heine16}. They introduce a toolbox of data descriptors (persistence
diagrams \cite{edelsbrunner02,  guillou_tvcg23},
merge
and contour trees
\cite{carr00,
gueunet_tpds19}, Reeb graphs \cite{biasotti08,gueunet_egpgv19}, or Morse-Smale 
complexes
\cite{robins_pami11, ShivashankarN12, gyulassy_vis18}), where
each descriptor captures specific structural relations between
the critical points  \cite{milnor63}
of the input scalar field. The
versatility of this toolbox has been demonstrated by the variety of domains
where successful applications have been documented, including
%
%
turbulent combustion \cite{bremer_tvcg11, gyulassy_ev14},
material sciences \cite{gyulassy_vis15, soler_ldav19},
fluid dynamics \cite{kasten_tvcg11, nauleau_ldav22},
bioimaging \cite{topoAngler},
or astrophysics \cite{shivashankar2016felix}. In particular, in
quantum chemistry, these methods enable a robust implementation of QTAIM
\cite{harshChemistry,ttk17, Malgorzata19} (with tailored 
molecular
descriptors \cite{chemistry_vis14}). At a conceptual level,
our work deals with the study of the variability of a
specific topological descriptor \revision{(the \emph{extremum graph}
\cite{CorreaLB11},
\autoref{sec_topology})}, within an ensemble of scalar fields, which is
a topic which also received a significant attention in recent years.
For instance, Athawale et al. \cite{athawale_tvcg19} introduce a framework for
the pointwise analysis of the \emph{geometrical} variability of separatrices in
the 2D Morse complex, based on an overlap-driven matching of \emph{mandatory
critical points} \cite{mandatory}.  Also,
general-purpose metrics \cite{NarayananTN15}
and tracking algorithms \cite{DasSN24} have been introduced
for extremum graphs. In contrast to the above methods, our work focuses on
studying the global \emph{structural} variability of separatrices in a specific
application
context (ensembles of 3D electron density fields), which comes with domain
specific hypotheses, which our work specifically exploits, resulting in a
simple and effective algorithm.

\ifdefined\includeSuggestions
\color{purple}

\subsection{Related work}
\label{sec_relatedWork}

\color{black}
\fi

\subsection{Contributions}
\label{sec_contributions}

This paper makes the following new contributions:
\begin{enumerate}
 \item \emph{Stability measure:} We present a stability measure, called
\emph{bond occurrence rate}, which associates to each bond path present at the
equilibrium state a
rate
of occurrence within an input ensemble of
electron densities. This measure enables the identification of
densities lacking certain bond paths, enabling detailed investigations.
 \item \emph{Algorithm:} We introduce an
 algorithm for the
computation of the above measure. It
exploits the specific hypotheses
of
molecular systems,
yielding a simple and effective
computation based on a tailored, geometry-aware partial isomorphism estimation
between the extremum graphs of the considered electron densities.
 \item \emph{Case studies:} We present two case studies, on rotational and
vibrational dynamic effects respectively, for
prominent
isomers
of
water hexamers (the so-called \emph{Ring}, \emph{Book}, \emph{Cage},
\emph{Prism}). Our stability measure enables the identification of
geometrical configurations favoring the disappearance of H-bond paths,
corroborating previously
documented observations. We also provide refined geometrical criteria for the
emergence of H-bond paths.
 \item \emph{Implementation:} We provide a C++ implementation of our approach
that can be used for reproducibility purposes.
 \item \emph{Database:} We contribute the database of electron densities
generated
for our case studies, i.e.,
4544 datasets (about 800 gigabytes),
organized in two ensembles of rotational effects (256 members each) and,
for each considered
isomer,
48 ensembles of vibrational effects (21 members
each). This database may constitute a benchmark for future research in ensemble
chemistry data.
\end{enumerate}

\section{Background}
\label{sec_background}

This section provides some technical background in chemistry and topology.
Specifically, it reviews the necessary notions of molecular chemistry with an
introduction to QTAIM
 \cite{matta07}.
Also, it presents a brief overview of the concepts
from Topological Data Analysis which are required for the robust
implementation of QTAIM. For further readings, we refer the reader to textbooks
on computational topology \cite{edelsbrunner09, zomorodianBook}.

\subsection{Hydrogen bonds}
\label{sec_background_Hbonds}
%

The term \emph{hydrogen bonds} (or H-bonds) in chemistry refers to subtle yet
essential attractive non-covalent interactions in molecular systems,
governing their structures, properties, and reactivity.

%

Formally, an H-bond 
is an interaction of the type Dn$-$H$\cdots$Ac,
where 
the \emph{donor} (noted Dn) and \emph{acceptor} (noted Ac)
typically refer to atoms of higher electronegativity
than the hydrogen atom (e.g., oxygen, nitrogen, fluorine).
For instance, in water, an O$-$H$\cdots$O hydrogen bond appears
between a 
hydrogen atom from one water molecule (with O$-$H and O in the role of H-bond donor) and
an oxgen atom of another molecule (H-bond acceptor).
From a quantum chemistry perspective, H-bonds appear as a subtle
combination of multiple phenomena such as charge transfer,
electrostatics and dispersion.
The specific balance of these effects depends on many factors, which explains
the challenges associated to the characterization of H-bonds.
Also, the bonding energies of H-bonds are in the range of $4$ to $40$
kJ/mol (much weaker than covalent bonds,
with bonding energies over 150 kJ/mol)
and they vary with the geometry of the system.
For these reasons (weak and
varying strengths), 
only generic guidelines are available in practice for characterizing the H-bond
appearance,
e.g.,
based on the H$\cdots$Ac distance
(in water, around $2$ Angstrom, about twice the O$-$H distance).

\begin{figure}
  \centering
  \includegraphics[width=0.495\linewidth]{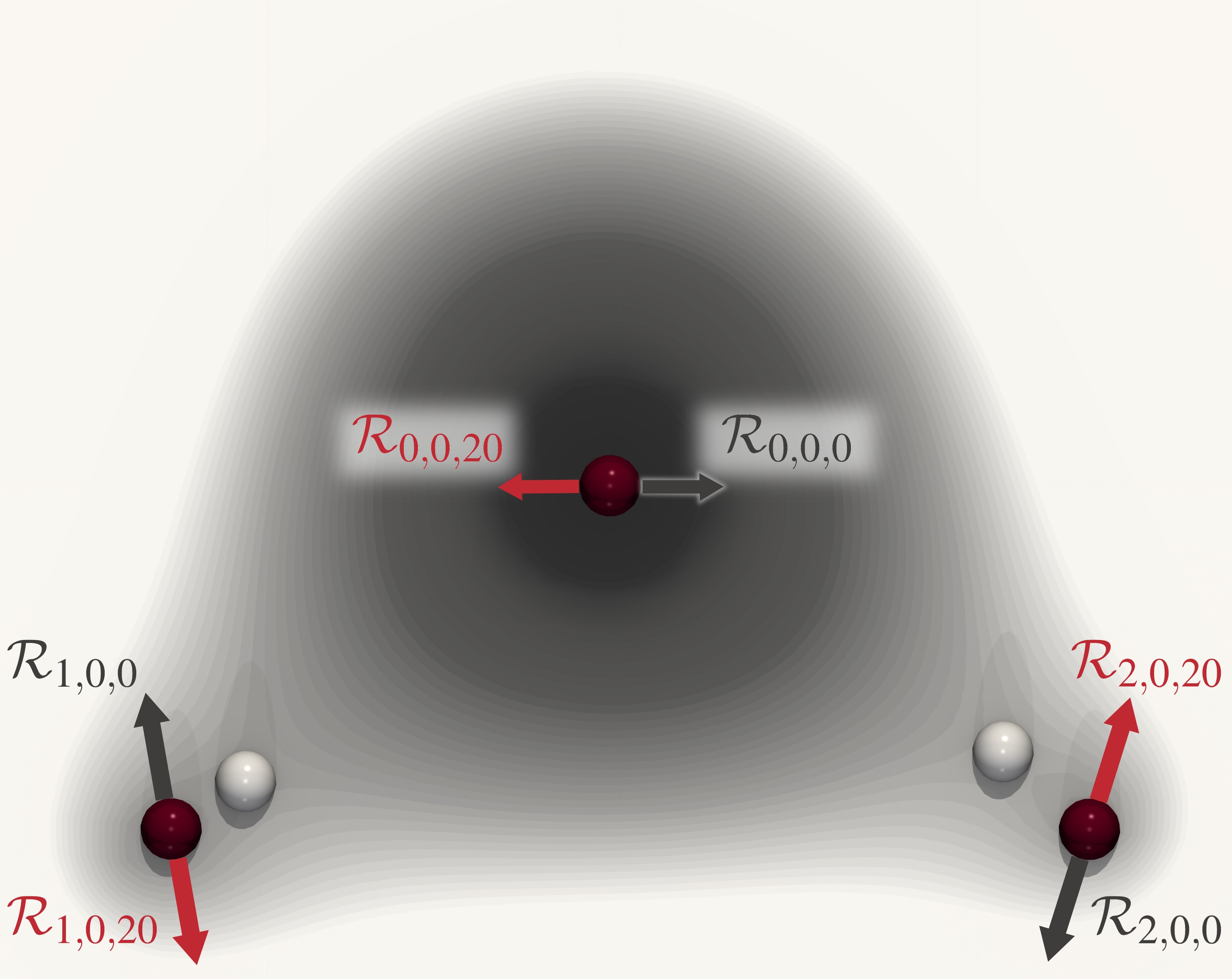}
  \hfill
  \includegraphics[width=0.495\linewidth]{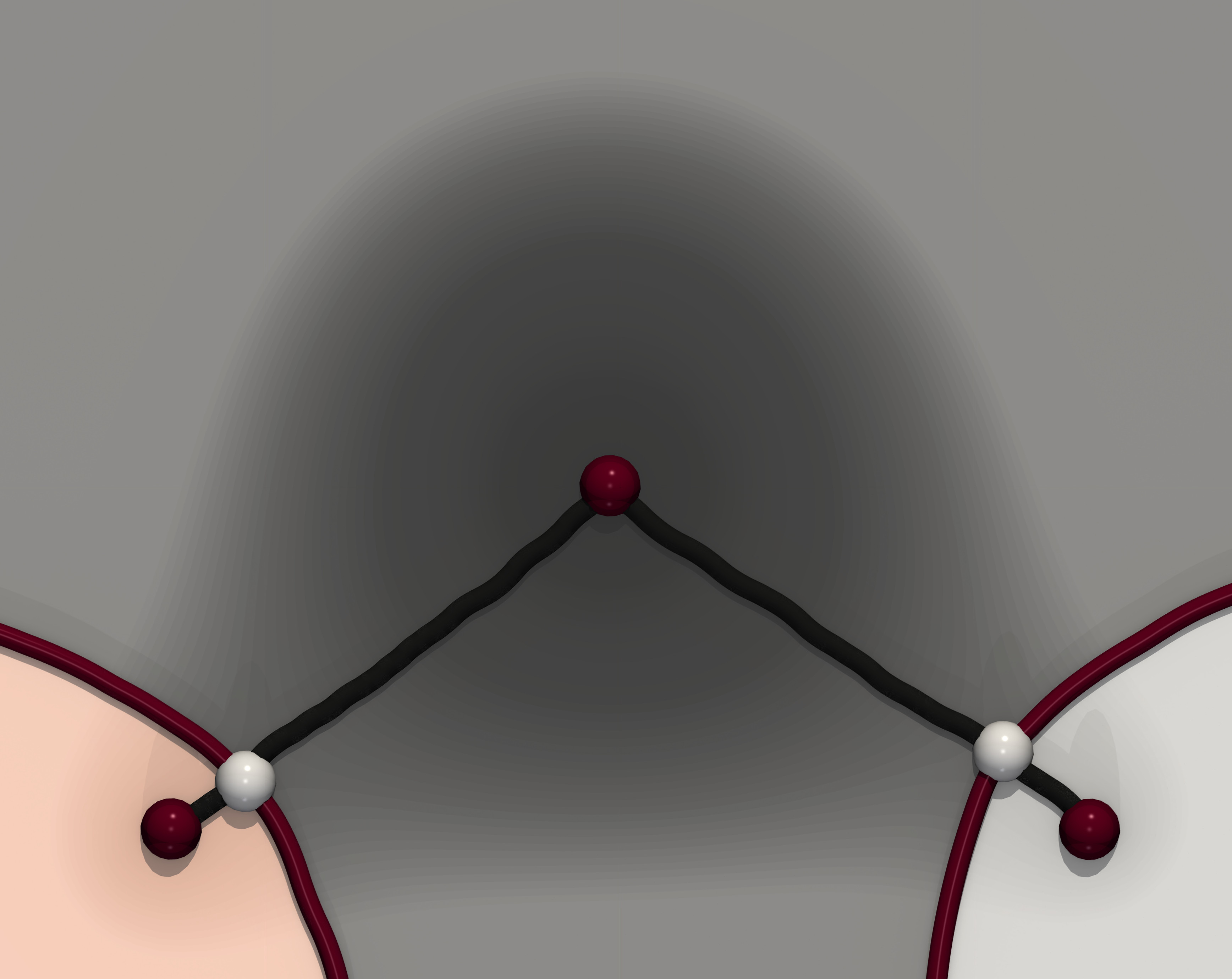}
  \caption{%
  Main concepts from the Quantum Theory of Atoms in
Molecule (QTAIM) on a 2D slice of the electron density for a water molecule.
\emph{Left:} Maxima (red spheres) of the electron density (gray gradient)
represent atoms, while saddles (white spheres) represent \emph{bond critical
points.} \revision{Arrows illustrate extreme vibrational displacement vectors
(mode $0$, see \autoref{sec_chemistry_vibrations})}.
\emph{Right:} Bond critical points
are connected to atoms
via \emph{bond paths} (black cylinders).
In QTAIM, two atoms that are considered chemically bonded are linked by a bond path
\cite{matta07}.
Each atom is associated with its \emph{basin} (colored regions, pink, light
gray and dark gray), whose boundary forms an \emph{interatomic surface} (red
curves on this 2D
\revision{slice)}.}
  \label{fig_qtaim}
\end{figure}

\begin{figure}
  \centering
  \includegraphics[width=.32\linewidth]{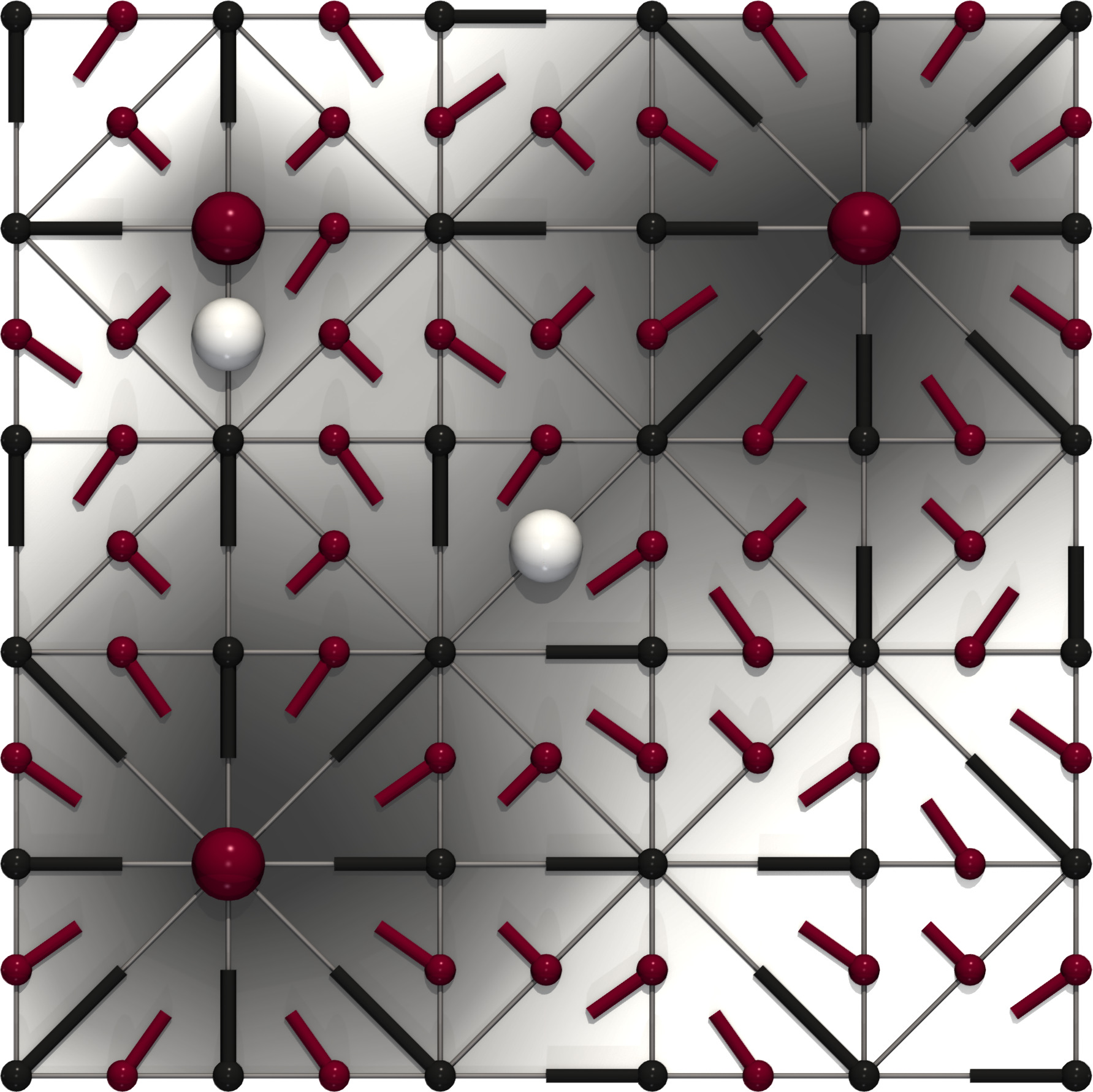}
  \hfill
  \includegraphics[width=.32\linewidth]{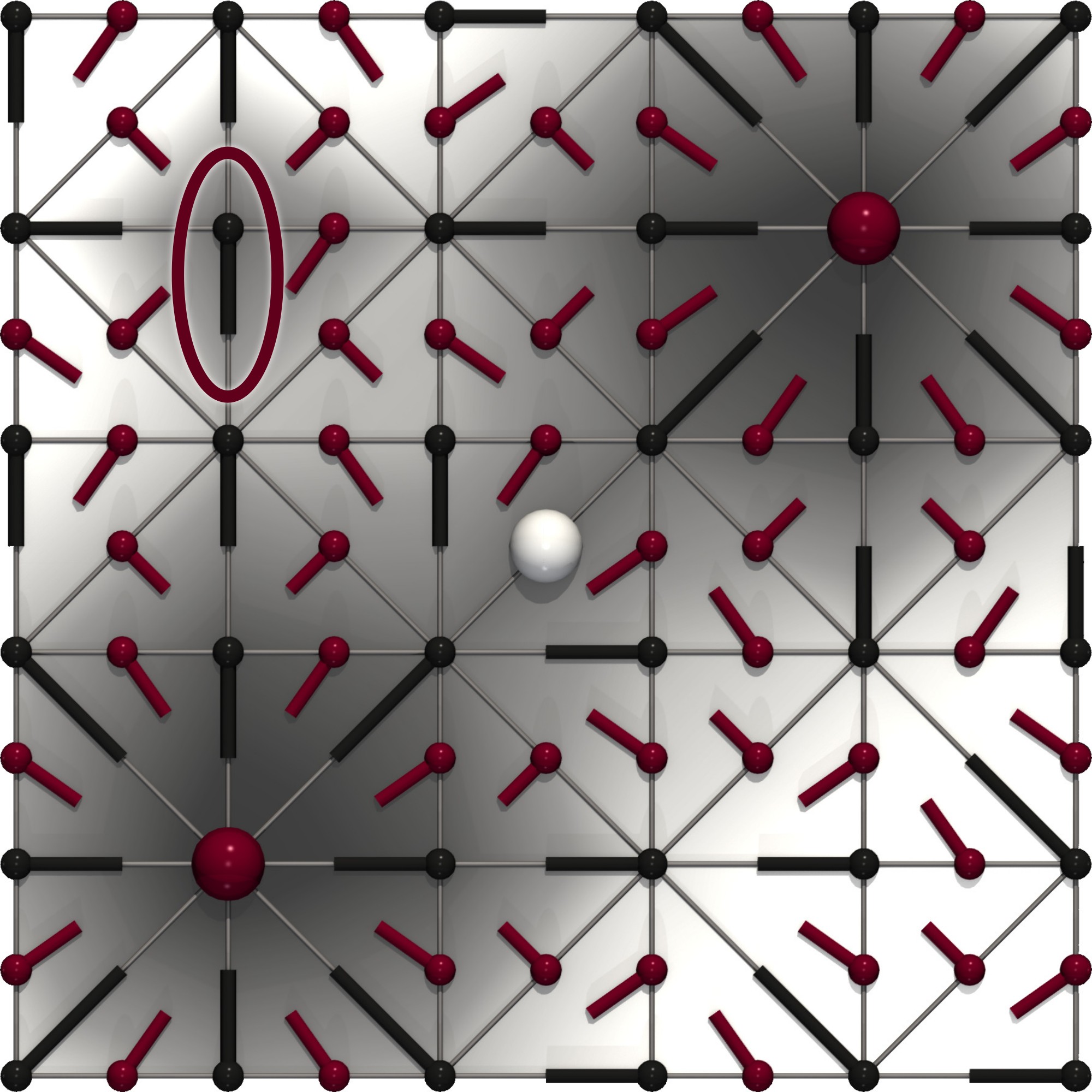}
  \hfill
  \includegraphics[width=.32\linewidth]{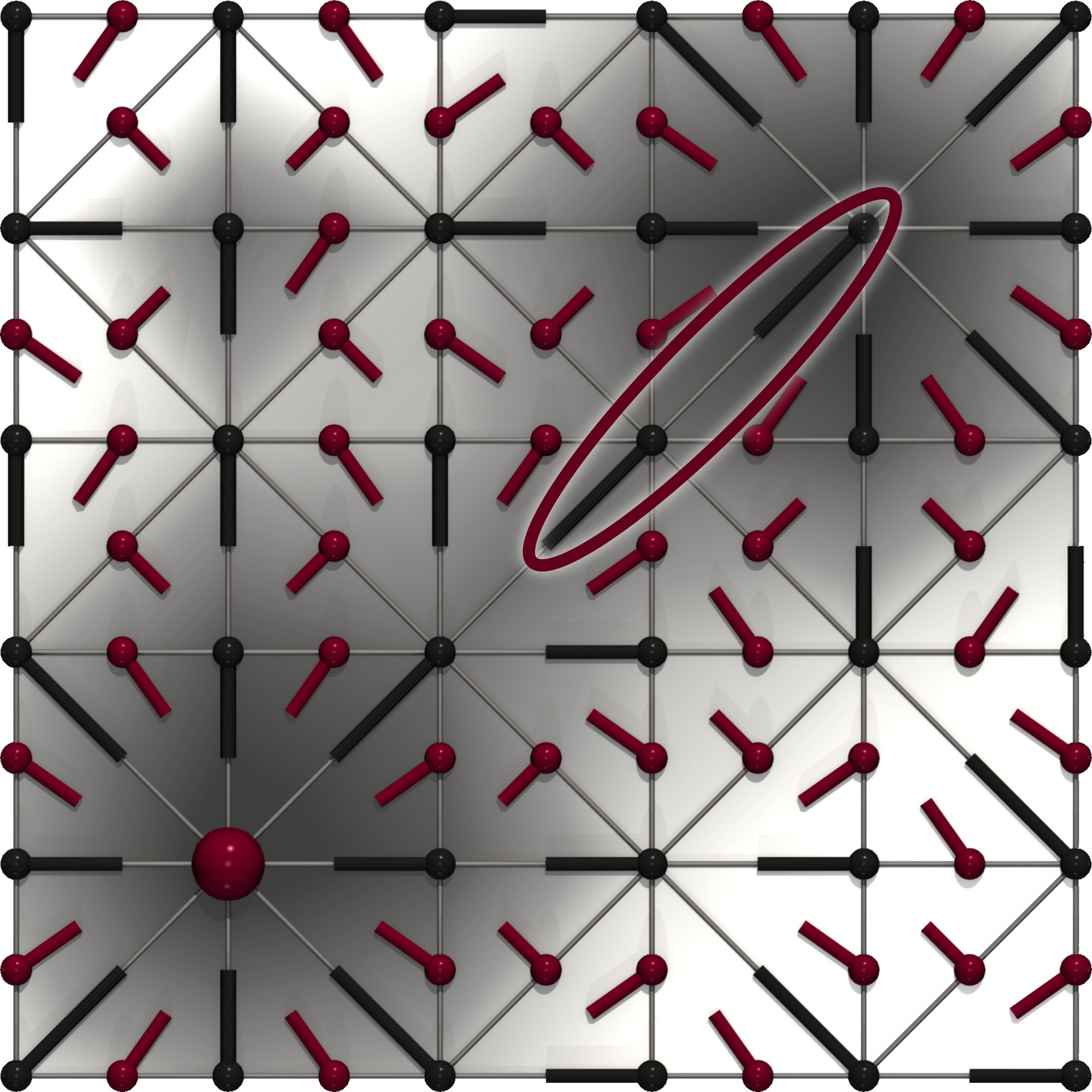}

  ~
  \vspace{-1ex}

  \includegraphics[width=.315\linewidth]{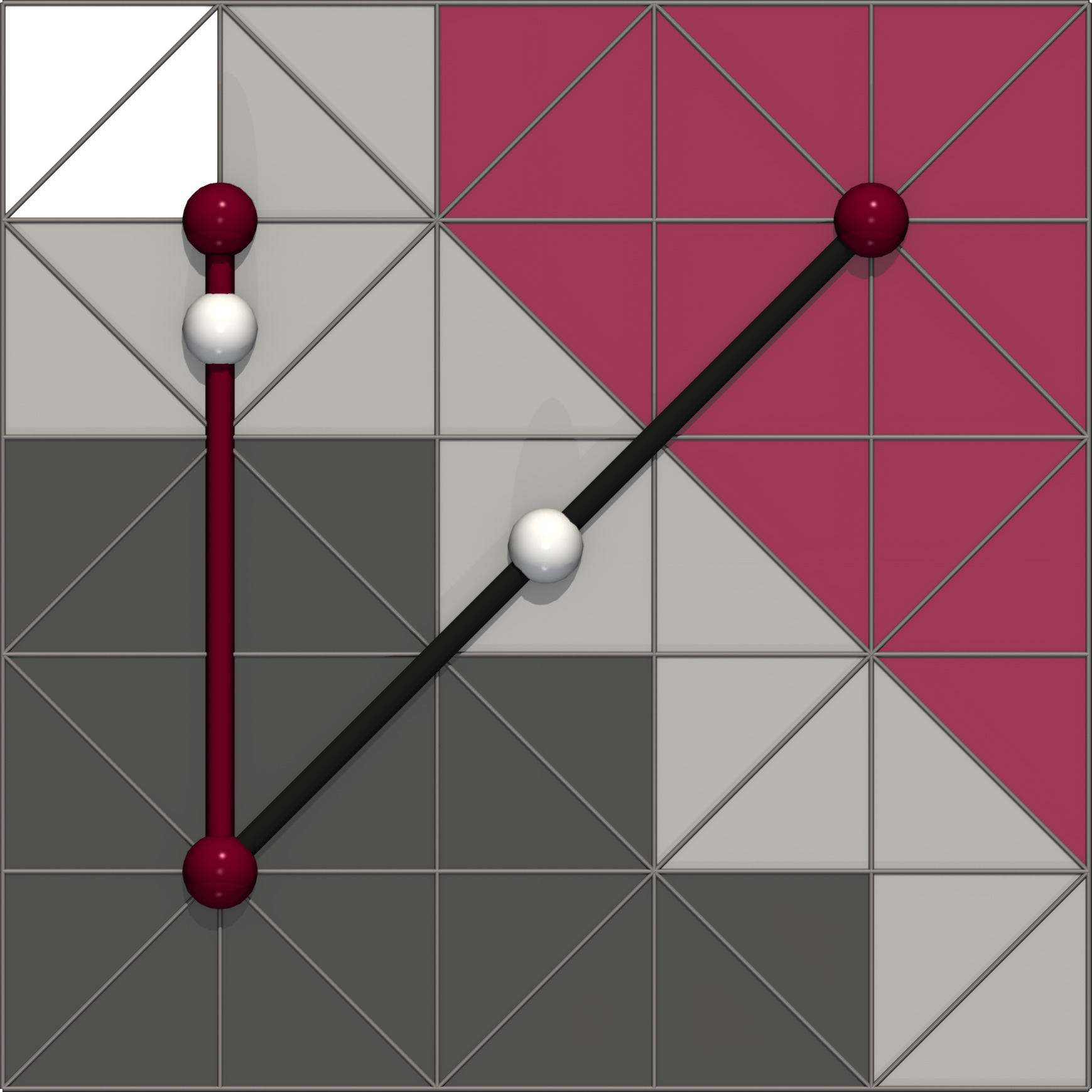}
    \hfill
  \includegraphics[width=.315\linewidth]{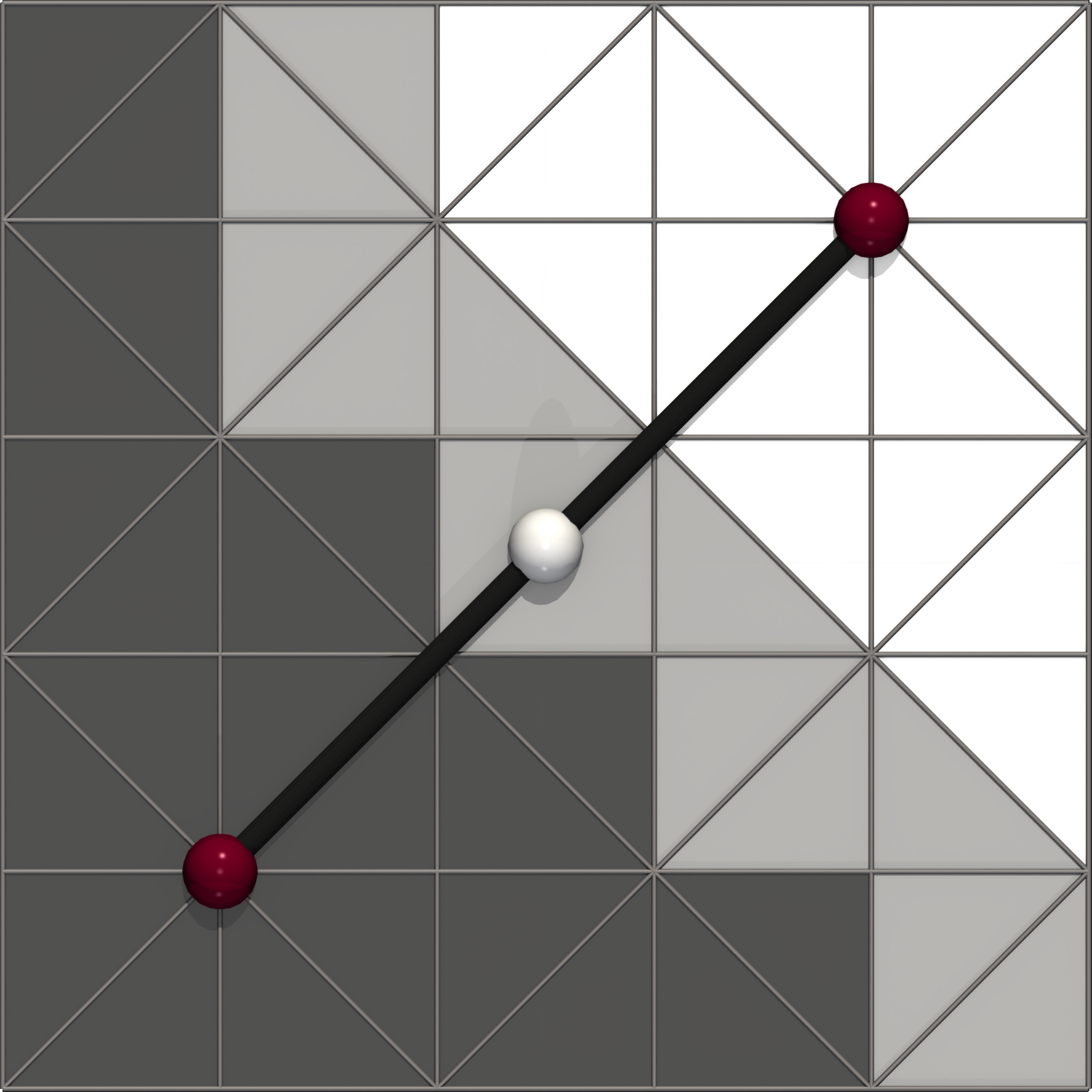}
    \hfill
  \includegraphics[width=.315\linewidth]{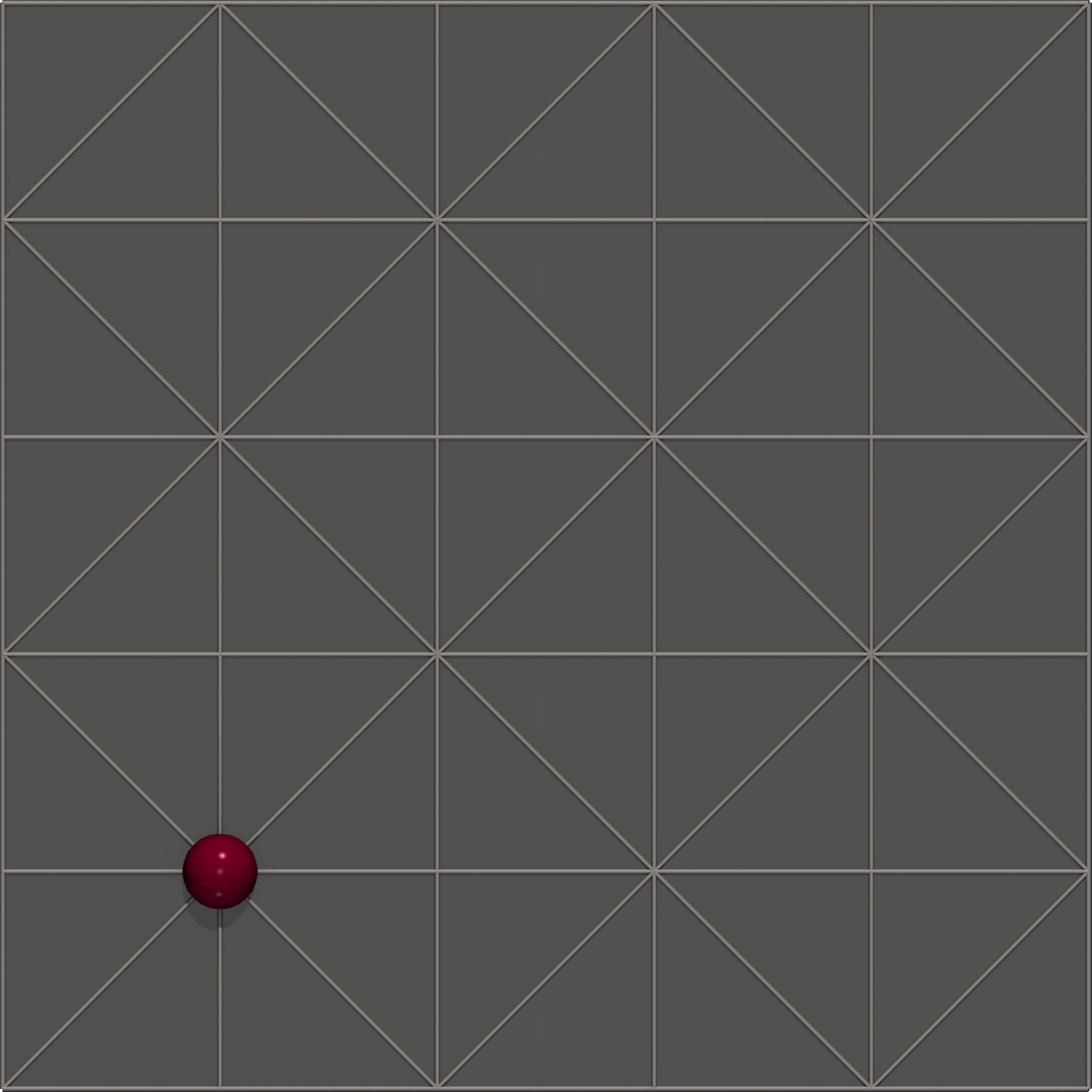}

  \caption{%
  Multi-scale Topological Data Analysis (from left to right) with discrete
Morse theory (DMT).
  \emph{Top:} Discrete gradient fields (black: vertex-edge vectors, red:
edge-triangle vectors) identify the simplices involved in no discrete vectors as
\emph{critical simplices} (red: minima, white: saddles). The field is
progressively simplified from left to right by \emph{v-path} reversal
(the vectors highlighted with an ellipse have been reversed).
  \emph{Bottom-Left:} The Morse complex decomposes the domain into \emph{stable
sets}.
  The white, dark gray and red regions only contain v-paths
involved in the stable sets of minima. The light gray region only
contains
v-paths
involved in the stable set of  saddles.
The \emph{unstable sets} of the saddles are reported with cylinders (red and
black).
As the gradient
field is progressively simplified (from left to right,
for instance by
\emph{topological persistence}
\cite{edelsbrunner02}), the resulting Morse complex is also
progressively simplified.
  }
  \label{fig_dmt}
\end{figure}

\subsection{Quantum Theory of Atoms in Molecules (QTAIM)}
\label{sec_background_qtaim}

Quantum Theory of Atoms in Molecules (QTAIM) was initially introduced by Bader
in the sixties and continuously developed since \cite{bader94, matta07}, as it
became a prominent methodology for analyzing and explaining specific phenomena
which proved challenging to explain with alternative models.
Drawing from the theory of gradient dynamical systems,
QTAIM relies on the
analysis of the electron density, noted $\electronDensity$, which, given a
molecular system, provides a three-dimensional distribution of the
electronic charge in the system (\autoref{sec_relatedWork}). Given the
specification of a molecular system,
this quantity can be computed based on density functional theory (DFT)
\cite{hohenberg64}. In particular, QTAIM provides a topological
specification
of the entities involved in molecular interactions, illustrated in
\autoref{fig_qtaim}. Local maxima
of $\electronDensity$ coincide with the nuclei of the atoms (red spheres,
\autoref{fig_qtaim}). The \emph{basin}
of a given atom A is defined as the region of 3D space for which all points
induce a gradient integral line (i.e., a curve everywhere tangential to the
gradient of $\electronDensity$, noted $\nabla \electronDensity$) converging to
A. 
The set of basins forms a
partition of the 3D space. Specifically, the
two-dimensional boundaries between basins form zero-flux surfaces (with
regard to $\nabla \electronDensity$), called \emph{interatomic
surfaces}. They describe interfaces in space
across which there is no net flow of 
electrons (red curves, \autoref{fig_qtaim}). Local maxima of $\electronDensity$
restricted to the interatomic
surfaces correspond to saddle points \cite{milnor63} of $\electronDensity$ in
3D, called \emph{bond critical points} \cite{matta07} (white spheres,
\autoref{fig_qtaim}).
Such points are the local minima of
\emph{bond paths} (black cylinders, \autoref{fig_qtaim}) connecting two nuclei,
which are formed by a pair of gradient integral lines started at
a saddle
and converging to the corresponding maxima.
According to QTAIM, if two atoms are chemically bonded (through a covalent
or non-covalent bond, such as an H-bond), they have their nuclei
linked by a bond path \cite{matta07} and the set of bond paths form the
molecular graph.


%

\subsection{Topological Data Analysis}
\label{sec_background_topology}

This section describes the necessary concepts from computational topology
\cite{edelsbrunner09, zomorodianBook} for a robust implementation of QTAIM in
practice.

\myparagraph{Input data:} The input data is typically provided by the DFT
computation as a scalar field $\rho : \cubicalComplex \rightarrow
\mathbb{R}$, valued on the vertices of a cubical complex $\cubicalComplex$
(i.e., a three-dimensional regular grid). In practice, for implementation
genericity purposes, $\cubicalComplex$ is often triangulated
(via an on-the-fly emulation \cite{ttk17}) into a simplicial complex $\domain$
(typically, with the Freudenthal triangulation \cite{freudenthal42}).
This yields a piecewise linear (PL) scalar field $\rho : \domain \rightarrow
\mathbb{R}$, valued on the vertices of $\domain$ and linearly interpolated on
the simplices of higher dimensions. At this stage, the input $\rho$ is assumed
to be injective on the vertices of $\domain$, which is enforced in
practice via symbolic perturbation \cite{edelsbrunner90}.


\myparagraph{Discrete gradient field:}
Given the above input formalization, one can leverage discrete Morse theory
(DMT)
\cite{forman98} to design robust, combinatorial \revision{algorithms} for
Topological Data
Analysis (as illustrated in \autoref{fig_dmt}). For this, the notion of
\revision{a}
discrete gradient field
can be introduced as
follows. Let $\{\simplex_i <
\simplex_{i+1}\}$ be a \emph{discrete vector}. It is formed by a pair of
simplices of $\domain$, such that $\simplex_i$ is an $i$-simplex and
$\simplex_{i+1}$ is a cofacet of $\simplex_i$ (see the vertex-edge vectors in
black and the edge-triangle vectors in red in \autoref{fig_dmt}, top).
$\simplex_{i+1}$ is usually referred to as the \emph{head} of the vector, while
$\simplex_i$ is its tail.
\revision{A}
\emph{discrete vector
field} on $\domain$ is defined as a collection of pairs $\{\simplex_i <
\simplex_{i+1}\}$, such that each simplex of $\domain$ is included in at most
one discrete vector. A simplex involved in no discrete vector is called a
\emph{critical simplex} (larger spheres in \autoref{fig_dmt}). The notion of
\revision{a}
\emph{v-path} introduces a discrete
analog to the concept of integral line. It is defined as a sequence of $k$
discrete vectors $\big\{\{\simplex_i^1 < \simplex_{i+1}^1\}, \dots,
\big\{\{\simplex_i^k <
\simplex_{i+1}^k\} \big\}$
such that \emph{(i)} $\simplex_i^j \neq \simplex_i^{j+1}$ (i.e., the tails of
two consecutive vectors are distinct) and \emph{(ii)} $\simplex_i^{j+1} <
\simplex_{i+1}^j$ (i.e., the tail of a vector in the sequence is a face of the
head
of the previous vector in the sequence) for any $1 < j < k$.
A v-path is said to terminate at a critical simplex $\simplex_i$ if $\simplex_i$
is a facet of the head of its last vector. By symmetry, a v-path is said to
start at a critical simplex $\sigma_{i+1}$ if $\sigma_{i+1}$
is a cofacet of the tail of its first vector.
\revision{We}
call the \emph{stable
set} of a critical simplex $\simplex_i$ the collection of v-paths which
terminate in  $\simplex_i$ (\autoref{fig_dmt}). By symmetry, we call the
\emph{unstable set} of
$\sigma_{i+1}$ the collection of v-paths which start in $\sigma_{i+1}$
(cylinders in \autoref{fig_dmt}, bottom).
Finally, a discrete vector field for which all v-paths are loop-free is called
a \emph{discrete gradient field}.

Several algorithms have been documented for
the computation of such a discrete gradient field from an input scalar field
\cite{robins_pami11, ShivashankarN12}.
\revision{In that context,}
the critical
simplices are
discrete analogs
to the
notion of
critical points from the smooth setting \cite{milnor63}, with their
dimension $i$ matching with the smooth notion of critical
\revision{indices.}

\myparagraph{Morse complex:} Given a discrete gradient field, the Morse complex
is the complex formed by the collection of stable sets (\autoref{fig_dmt},
bottom). In practice, to cope
with noise or numerical inaccuracies, the Morse complex is often simplified,
either in a post-process \cite{GyulassyBHP11} or in a
pre-process \cite{GuntherRSW14}, by modifying the discrete gradient field prior
to computing the Morse complex. In particular, a pair of critical simplices
$\simplex_i$ and $\simplex_{i+1}$ can be simplified together if there exists a
unique v-path connecting them, by simply reversing the discrete vectors along
the v-path \cite{forman98}.
\revision{Globally,}
the Morse complex can be
progressively simplified, by iteratively reverting v-paths connecting pairs of
critical simplices (\autoref{fig_dmt}, from left to right).
In practice, several criteria can be considered for selecting a v-path for
reversal. For instance, \emph{topological persistence} \cite{edelsbrunner02,
edelsbrunner09} is an established, general-purpose importance measure, that is
well studied in Topological Data Analysis.
Overall, this combinatorial
approach enables a robust and
straightforward implementation of QTAIM \cite{harshChemistry, ttk17,
Malgorzata19, ttk19}.
%
%
Specifically, since
in DMT
discrete vectors point down
\cite{forman98}, the QTAIM \emph{basins} correspond to the stable sets of the
minima of the opposite of the electron density, \revision{and the
\emph{interatomic
surfaces} correspond to the stable sets of its $1$-saddles.}
%
\revision{Also,}
the \emph{bond
paths} are given by the unstable sets of $1$-saddles of this opposite density.
%
%
%


\ifdefined\includeSuggestions
\color{purple}

\color{black}
\fi

\section{\gosia{Electron Density Ensemble Generation}}
\label{sec_chemistry}

This section describes the generation of the ensemble datasets considered in
our case studies (\autoref{sec_useCase_pathways}
and \autoref{sec_useCase_vibrations}).
\revision{These ensembles model, for selected water clusters
(\autoref{sec_chemistry_configuration_selection}),  \emph{proton tunneling}
(\autoref{sec_chemistry_pathways}) and \emph{molecular vibrations}
(\autoref{sec_chemistry_vibrations}). These are two important
phenomena naturally occurring in chemical processes,
which can trigger
\emph{isomerization} --
transforming a molecule into
another, with identical composition, but with a distinct
spatial arrangement of atoms, resulting in
distinct chemical properties. Isomerization plays
a crucial role in many areas of chemistry (e.g., drug
design) and understanding the conditions favoring its occurrence is an important
challenge in
quantum
chemistry.}


\noindent

\begin{figure*}
\centering
%
\includegraphics[width=\linewidth]{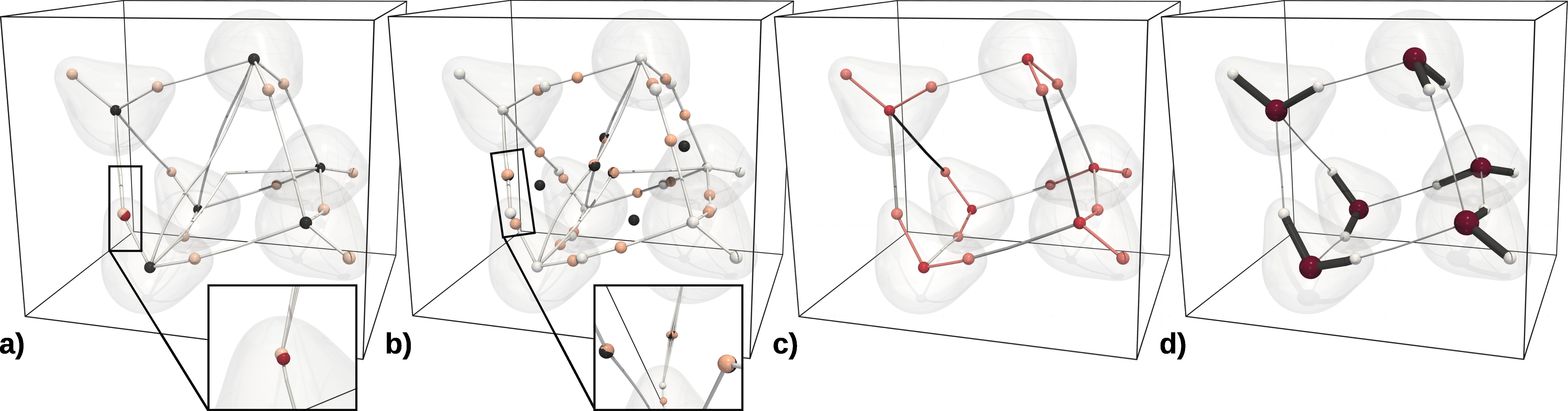}

\caption{%
Extremum graphs
of
\revision{opposite}
electron density
\revision{($\rho' \in (-126,0)$)},
at various stages
of \emph{bond graph} extraction, for
a
\emph{Prism} water hexamer (transparent: isosurface for
$\rho'=-0.1$).
\emph{(a)} The presence of spurious minima (red sphere) requires
topological simplification \cite{Lukasczyk_vis20}. Minima are colored
by
\emph{topological persistence}
(\autoref{sec_background_topology}).
In this example,
the spurious
minimum (red) has a persistence of $4 \times 10^{-4}$, while the first
non-spurious one has a persistence of $3.6 \times 10^{-2}$ (two order of
magnitude gap).
\emph{(b)} The presence of spurious pairs of $1$ and $2$-saddles (orange and
black spheres respectively) requires a step of simplification, achieved by
v-path reversals, ordered by function value difference.
\revision{Here,}
the
last spurious saddle-saddle v-path exhibits a function difference of $4 \times
10^{-5}$, while the first non-spurious one has a difference of $5 \times
10^{-3}$ (two order of magnitude gap).
\emph{(c)} The simplified extremum graph is represented with its nodes and arcs
 colored by $\rho'$ (nodes: value at the minimum, arcs: value
at the saddle), enabling a clear distinction between oxygen  (red) and
hydrogen (pink) atoms  as well as between covalent (pink) and non-covalent
(\revision{white}
to black) bonds.
\emph{(d)} The resulting bond graph represents the identified atoms (red:
oxygen, white: hydrogen) and bond types (dark gray: covalent, light gray:
H-bond).
}
\label{fig_analysisPipeline}
\end{figure*}

\subsection{\gosia{Water cluster selection}}
\label{sec_chemistry_configuration_selection}

\julien{%
The study of water clusters, (H$_2$O)$_{n_m}$, is essential for the
understanding of the structure of liquid water and ice
\cite{yoo.xantheas_hocc_2017,gao.etal_sr_2022}.
It provides key insights into the mechanisms of many
processes, such as water aggregation, microsolvation
\cite{santis.etal_tjocp_2024,hadad.etal_ijqc_2019,zhang.xu_a_2022}
or proton transfer \cite{wolke.etal_s_2016}.}

\julien{For a given number of molecules $n_m$, water clusters can exhibit
distinct
isomeric forms (i.e.,
geometrical arrangements), as each water
molecule can partcipate in four H-bonds.
Each
isomer
has a specific \emph{hydrogen bond network} (i.e., set of
H-bonds), resulting in unique chemical properties.
Moreover, isomers with
comparable energies can easily rearrange into one another through a process
called \emph{interconversion} \cite{gao.etal_jcs_2024}.}
\revision{In our work, we focus on these
relatively
simple yet chemically
important systems
due to
their extensive documentation and ubiquity in chemistry.
For example, several studies document
proton tunneling (\autoref{sec_chemistry_pathways}) or
molecular vibration (\autoref{sec_chemistry_vibrations}) for these clusters,
hence helping
validating our approach. Also, their geometric scale,
between micro- and macro-systems,
typically
makes them good candidates in  chemistry
for investigating novel
models or experimental protocoles, before
considering more complex systems (see \autoref{sec_conclusion}).}

\revision{Specifically, we focus on water}
\julien{hexamers ($n_m = 6$), \revision{which} are prominent examples of
water
clusters
\cite{gao.etal_sr_2022}, as they constitute
the smallest three-dimensional cluster whose non-quasiplanar isomers (the
\emph{Book}, \emph{Cage} and \emph{Prism}) exhibit lower energies
than quasiplanar isomers (such as the \emph{Ring}). For this reason,
(H$_2$O)$_{6}$ is often referred to in the literature as \emph{``the
smallest drop of water''}
\cite{nauta.miller_s_2000}.}
\julien{High resolution spectroscopy
as well as state-of-the-art quantum
calculations \cite{saykally.wales_s_2012,perez.etal_s_2012}
indicate that the \emph{Prism}, the \emph{Cage} and the \emph{Book} are the
isomers of
(H$_2$O)$_{6}$
with the lowest energies.
\revision{Consequently,} they likely coexist, even at low
temperatures.
\revision{Therefore,}
we focus our study on
these three isomers as well as the \emph{Ring} isomer, which is the quasiplanar
structure with the lowest energy.}

\subsection{\gosia{Tunneling pathways}}
\label{sec_chemistry_pathways}
\julien{To evaluate the chemical relevance of our approach, we first study the
\emph{Prism} isomer
under the effect
of \emph{proton tunneling} -
a phenomenon in which a molecule in its equilibrium
state rearranges to an equivalent (i.e., isoenergetic) structure, for example
by permuting equivalent atoms \cite{quack.seyfang_msaqd_2021}.
Proton tunneling in the
\emph{Prism}
isomer of (H$_2$O)$_{6}$ was observed experimentally
and confirmed by calculations
\cite{rotations16,wang.etal_nr_2016,vibrations1}.
Specifically,
far-infrared vibration-rotation tunneling (FIR-VRT)
spectroscopy observations report characteristic
splitting patterns.
These
can be explained by two
distinct
tunneling mechanisms, associated to specific rotations of water molecules:
\emph{(i)} An \emph{anti-geared} rotation,
breaking one H-bond, and
\emph{(ii)} a \emph{geared} motion, simultaneously breaking two H-bonds.
Each of these two motions can be modeled by a specific \emph{pathway} (i.e., a
temporal sequence of transitory
structures between the two equivalent states), which can be estimated with 
the ring-polymer molecular dynamics (RPMD)
\revision{approach} \cite{richardson_tjocp_2018}.
For each step $\alpha_k \in (0, 255)$ of the pathways
reported in \cite{rotations16}, the  electron density was computed
with a non-relativistic DFT method \cite{hohenberg64}, with the
PBE0
functional \cite{adamo.barone_tjocp_1999} and the TZ2P basis, as implemented in
the ADF software \cite{tevelde.etal_jcc_2001} and 
facilitated through PyADF\cite{focke.etal_jcp_2024}. The
computations were carried
\revision{out}
on a $256^3$ regular grid, with a \revision{sample} spacing of $0.05$
Angstrom (\r{A}).}

\subsection{\gosia{Molecular vibrations}}
\label{sec_chemistry_vibrations}

\julien{%
\emph{Molecular vibrations} are intrinsic behaviors of molecular systems.
They describe the relative motion  of atoms (occurring at all temperatures)
which does not change the position of the molecular center of mass.
\revision{In the following, although our
molecular vibration modeling
follows the typical approach in chemistry \cite{wilson_tjocp_1939},
we still specify its details in order to precisely
document
our data generation process,
thereby enabling reproducibility.}
A system composed of $n_a$ atoms
is a subject to $n_m = 3n_a - k$ vibrational degrees of freedom, with $k
= 3 + 3$ in non-linear molecules (to discard the 3 translational and  3
rotational degrees of freedom of the molecular center of mass). In particular,
for a water hexamer, (H$_2$O)$_{6}$, $n_a= 18$ and 
$n_m = 48$. The complex motion induced by these degrees of freedom can be
decomposed into simpler components,
referred to as the \emph{normal modes of vibrations}.
 Each mode describes the
motion of all atoms occurring with a specific frequency and phase.
}

\julien{%
\revision{The}
equilibrium structure of each isomer, noted ${\mathcal R}^0$, was
optimized with the PBE0 TZ2P model (\autoref{sec_chemistry_pathways}) and the
harmonic vibrational modes, noted
${\mathcal Q}$,
were computed via
normal mode analysis (NMA), a standard method
for studying
molecular vibrations under the harmonic oscillator approximation
\cite{wilson_tjocp_1939}.
\revision{The}
displacement of a nucleus
$i$
along each of the $n_m$ vibrational normal vectors
${\mathcal Q}_j \in \mathbb{R}^{3 n_a}$
is \revision{then}
given by:
\begin{equation}
\nonumber
    {\Delta}_{i,j} = s_i {\mathcal L}_{i,j} {\mathcal Q}_j,
    \qquad    
    i \in \{1,\ldots,n_a\},
    \label{eq:normal-to-Cartesian_coords}
\end{equation}
where $s_i$ is a coefficient accounting for the masses of the oscillating
nuclei and where $\mathcal{L}_{i,j}$ is constructed
from the second-order (Cartesian) derivatives
of the potential energy
at the equilibrium atomic positions.}

\julien{%
Next, let $\alpha_k$ be
an integer in the range $(0, 20)$\revision{, which constitutes a sufficient
sampling for our case study (to enable the
observation of clear transitions in H-bond configurations,
see \autoref{sec_useCase_vibrations})}.
\revision{The}
position of a nucleus $i$ in the perturbed
geometry $k$ of the vibration mode $j$ is:
\begin{eqnarray}
\nonumber
\mathcal{R}_{i,j,k} = \mathcal{R}^0_i + 0.4 \cdot (\alpha_k-10) \cdot
\mathcal{A}_h \cdot \Delta_{i,j},
\end{eqnarray}
where $\mathcal{A}_h$ is a classical estimation of the \emph{turning
point} of the harmonic oscillatory motion \cite{goldstein.etal__2016}.}
\revision{\autoref{fig_qtaim}
illustrates the resulting vectors on a
water molecule at mode $j=0$, for extreme displacements (i.e., $k \in \{0,
20\}$).}

\julien{%
Finally, given the obtained out-of-equilibrium structures $\mathcal{R}_{j,k}$,
we  considered $n_m = 48$ ensembles of electron
densities for each selected isomer
(\autoref{sec_chemistry_configuration_selection}), calculated with the
the same computational model as used in \autoref{sec_chemistry_pathways}.
Each ensemble describes one vibrational normal mode and
counts $n_v = 21$ electron densities generated from:
\emph{(i)} the isomer at equilibrium ($\alpha_{10}$) and
\emph{(ii)} its $20$ derived geometries, obtained by displacing each atom along
its vibrational normal vector ($10$ in negative directions $\alpha_k
\in \{0, 9\}$, and $10$ in positive
directions $\alpha_k \in \{11, 20\}$).
}

\ifdefined\includeSuggestions
\color{purple}

\color{black}
\fi



\section{Bond Graph Extraction}
\label{sec_topology}

This section documents the topological analysis of a single electron density
dataset, for the automatic extraction of a graph, which we call \emph{bond
graph}, representing the covalent and non-covalent bonds in the
system.
While the pipeline described in this section is typical of related work
\cite{harshChemistry, Malgorzata19}, we emphasize here the
details that are specific to both water hexamers and the analysis of the
stability of their hydrogen bonds.

\subsection{Chemical features of interest specification}
\label{sec_chemical_features}

Given the electron density
$\rho: \domain \rightarrow \range$
of a water
hexamer,
\revision{and}
following QTAIM (detailed in \autoref{sec_background_qtaim}),
the goal of our
analysis is to extract a
geometrical representation of the following features of interest:
oxygen atoms, hydrogen atoms, covalent bonds, H-bonds.

As described in \autoref{sec_background_topology}, since in discrete Morse
theory discrete vectors are pointing down, we will consider in the following
the \emph{opposite electron density}, i.e., the field $\rho' : \domain
\rightarrow \range$, such that $\rho'(v) = - \rho(v)$ for all the vertices $v
\in \domain$.
Then, the nucleus of each atom can be identified as a local
minimum of $\rho'(v)$.
Within each molecule, a \emph{covalent bond}
can be
extracted as
the \emph{bond path} (i.e.,
the unstable set, \autoref{sec_background_topology}) of a
\emph{bond critical point} (i.e., a $1$-saddle of $\rho'(v)$ located within a
molecule).
Finally, \emph{H-bonds} can be extracted as the remaining \emph{bond paths}
which connect a hydrogen atom to an oxygen atom.
We describe in the following two sections how to robustly extract these
features.


%
%

\subsection{Topological characterization}
\label{sec_topological_features}
This section documents the topological analysis of the input opposite
electron density field $\rho' : \domain \rightarrow \range$.

Scalar data is often impacted by noise, even
with
smooth data such as
the electron density, as shown in
\autoref{fig_analysisPipeline}(a), where spurious minima occur in $\rho'$,
due to local numerical inaccuracies. To address this issue, we pre-process
$\rho'$ with persistence-driven simplification.
\julien{Specifically, the minima associated with a measure of \emph{topological
persistence} \cite{edelsbrunner02, edelsbrunner09} lower than
a threshold $\epsilon$
are simplified with local perturbations of the scalar data, using
the
algorithm presented by Lukasczyk et al. \cite{Lukasczyk_vis20}.
In particular, $\epsilon$ is selected such that only $18$ minima remain in the
field ($3$ atoms per water
molecule, for each of the $6$ molecules of the system). In practice,
$1 \times 10^{-3}$ is a typical cut-value for $\epsilon$
(see \autoref{fig_analysisPipeline}).}

\begin{figure}
\centering

\includegraphics[width=\linewidth]{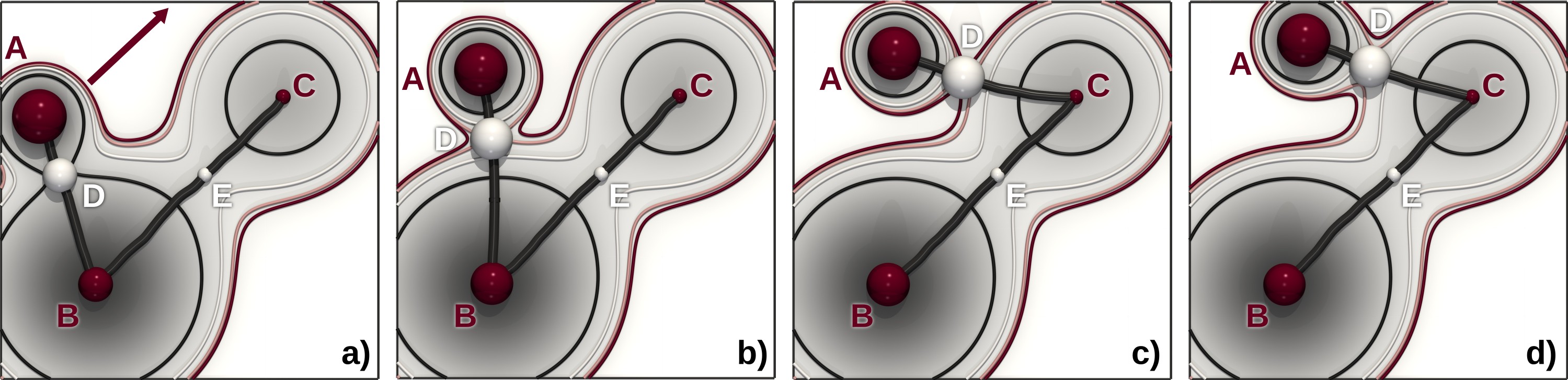}
\caption{Instability of
unstable sets on a 2D Gaussian mixture (with negative coefficients, gray
gradient; four isolines are shown, from black to red). From \emph{(a)} to
\emph{(d)}, the
minimum $A$ (red sphere) is translated along a direction parallel to the axis
$B-C$ (red arrow). From \emph{(b)} to \emph{(c)}, only a slight displacement
of $A$ is sufficient to switch the position of the saddle $D$ (white sphere),
resulting in a drastic change in the structure of the unstable sets: $A$ is now
connected to $C$ instead of $B$.
Note that this instability is not necessarily correlated to \emph{topological
persistence}
(radius of the spheres, see \autoref{sec_background_topology}), as
the saddle $D$ is involved in a persistence pair more persistent than that
of $E$. This example further motivates our investigation of the practical
stability of QTAIM bond-paths.
}
\label{fig_unstability}
\end{figure}

Similarly, numerical inaccuracies can also generate spurious pairs of saddles
in the input field $\rho'$ (\autoref{fig_analysisPipeline}(b)). As described in
\autoref{sec_background_topology},
these can be removed via iterated v-path reversals. Specifically, we revert a
v-path connecting a $2$-saddle down to a $1$-saddle if their function value
difference is  less than \julien{$1 \times 10^{-4}$}.

Once the above two pre-processes are completed (to remove spurious minima and
$1$-saddles), the \revision{\emph{extremum graph}}
\cite{CorreaLB11}
is computed by collecting the
\emph{unstable sets} (\autoref{sec_background_topology}) of each remaining
$1$-saddle.
Formally, the extremum graph of $\rho'$, noted $\extremumGraph(\rho')$, is a
graph with a set of nodes $\minimumSet(\rho')$ and a set of arcs
$\unstableSets(\rho')$, such that each node $n \in \minimumSet(\rho')$
represents exactly one minimum of $\rho'$, and each arc $a \in
\unstableSets(\rho')$ represents the unstable set of a $1$-saddle of $\rho'$
linking two of its minima.


Note that, for a reliable structural analysis of the extremum
graph, the geometrical accuracy of the unstable sets is of paramount
importance. For that, among the different discrete gradient
algorithms studied in the literature \cite{robins_pami11,
ShivashankarN12, empiricallyConvergent, GyulassyBP12,
gyulassy_vis18, ThanhAW24}, we use the approach by Gyulassy et al.
\cite{GyulassyBP12}, as it provides improved
accuracy and provable
convergence to the underlying continuous flow (for its stochastic variant).

Interestingly,
from a mathematical point of view,
as their name suggests, unstable sets are fundamentally unstable
constructions, even when considering geometrically accurate discrete gradient
fields.
\autoref{fig_unstability} shows an example of a 2D scalar field, where an
arbitrarily
small perturbation results in a
drastically
different
extremum graph.
Despite this
instability, unstable sets are
widely used in QTAIM for characterizing covalent and non-covalent bond paths
(\autoref{sec_background_qtaim}).
This paradox further motivates our
ensemble-based
H-bond stability analysis.

\subsection{From topological to chemical features}
\label{sec_topology2chemistry}
This section documents the extraction of the \emph{bond graph}
from the extremum graph computed in the previous section.

Since oxygen and hydrogen atoms have notably different atomic masses, these
can be easily distinguished based on their opposite electron density values
($\rho'$), typically
in our database
around
\julien{$-8$}
for oxygen atoms and above $-1$ for hydrogen
atoms.
%
%
\revision{Thus, we classify as oxygen atoms all the atoms with an opposite
electron density value below $-4$, and as hydrogen atoms all the atoms above
this value.}

Similarly, since they involve strong electrostatic interactions, covalent bonds
can be easily distinguished from non-covalent bonds based on the opposite
electron density value ($\rho'$)  of their corresponding bond critical point.
%
%
\revision{Specifically, we classify as covalent bond an unstable set from the
extremum graph which connects an oxygen atom to a hydrogen atom, and whose bond
critical point has an opposite electron density value
below $-0.1$. Symmetrically, unstable sets connecting an oxygen atom to a
hydrogen atom, and whose bond critical point has an opposite electron density
value
above $-0.1$ are classified as non-covalent bonds.}
Note that this isovalue of opposite
electron density ($-0.1$)
is often selected for
representing the geometry of a molecule
via isosurfacing 
(\autoref{fig_analysisPipeline}).
In certain
cases,
unstable sets may connect two oxygen atoms, which
does
not characterize a priori any chemically relevant interaction
in this context. We
will refer to such
connections as \emph{misconnected H-bonds} (see \autoref{sec_discussion}).

Finally, the \emph{bond graph} is obtained by collecting the
covalent and hydrogen bonds identified above,
see \autoref{fig_analysisPipeline}(d). This graph will be later exploited in
our use cases (\autoref{sec_results}) to evaluate bond stability.


%
%

\begin{figure}
  \centering
%
%
%

  \includegraphics[width=\linewidth]{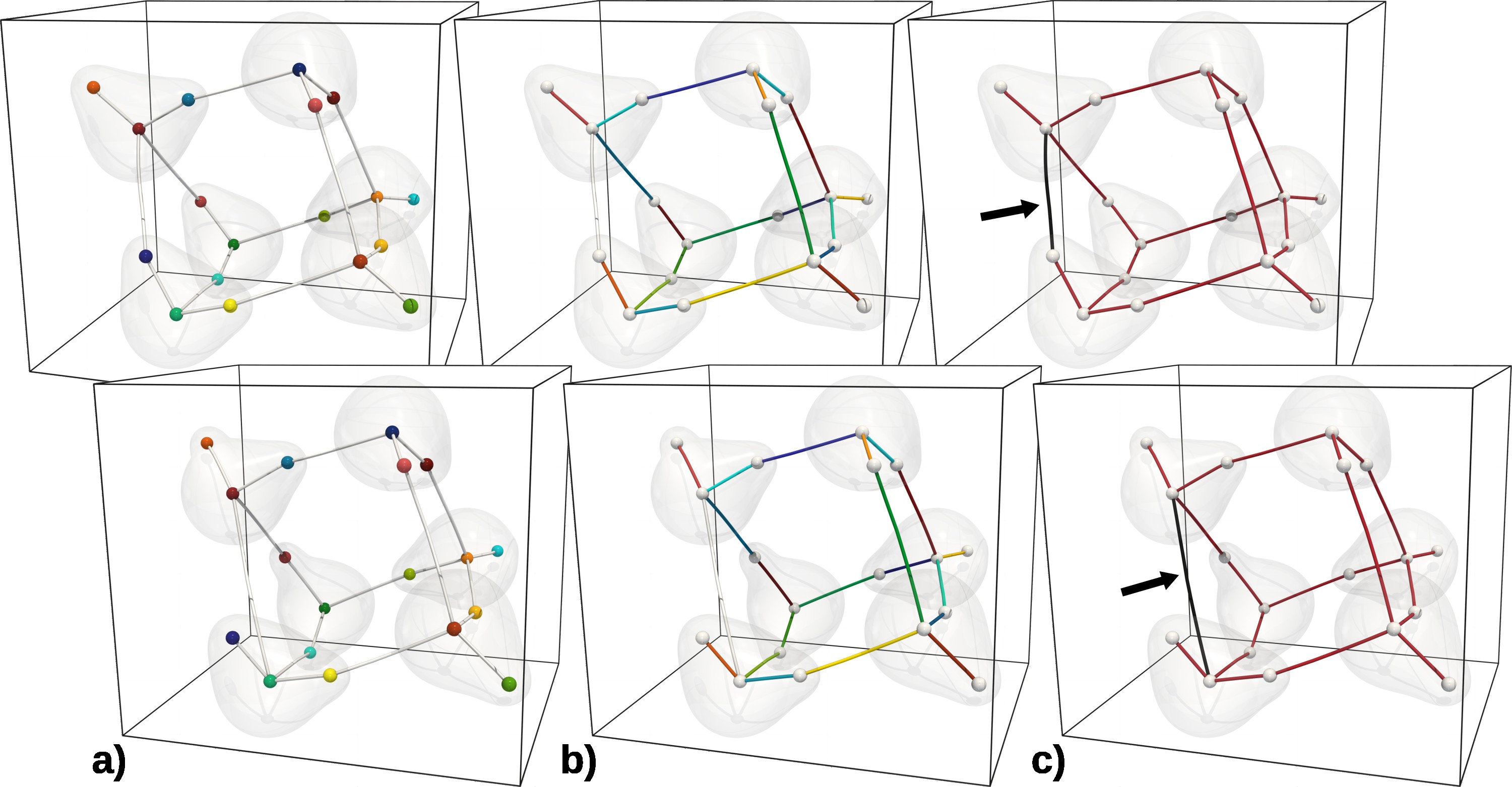}

%
%

%
%
%
%

  \caption{Geometry-aware partial isomorphism estimation between the extremum
graphs of the \emph{Prism} water hexamer at equilibrium (top) and under large
vibrational displacement (bottom).
\emph{(a)} First, an optimal assignment
matches  the nodes together (colors), in order to minimize the sum of their 3D
\revision{distances}
(\autoref{eq_energy}).
\emph{(b)} This matching is extended to the
arcs (colors) by matching arcs together if their nodes are matched by the
optimal node assignment, possibly yielding a partial isomorphism (unmatched
arcs are shown in white).
\emph{(c)} For this ensemble (only two graphs),
the \emph{bond occurrence rate} reports a value of $1$ (red) for all arcs
involved in the partial isomorphism and of $0.5$ (black) for the others.}
  \label{fig_partialIsomorphism}
\end{figure}

\begin{figure*}
\centering
\includegraphics[width=\linewidth]{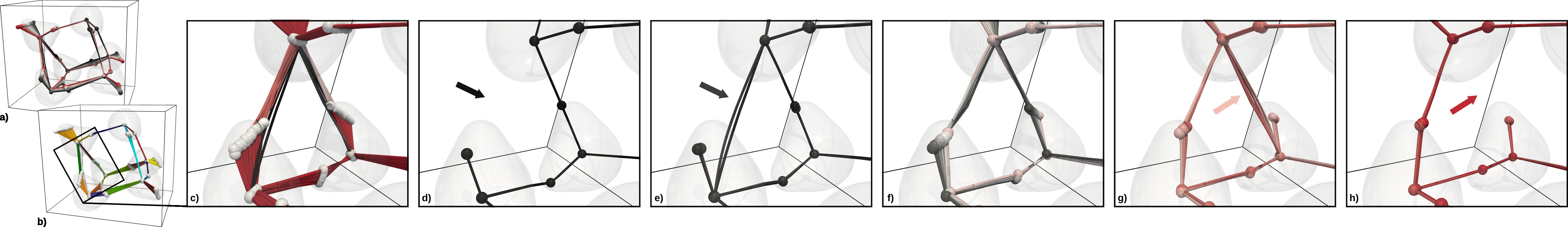}
\caption{Bond occurrence analysis for
the
\emph{Prism} (vibration mode $0$).
\emph{(a)} Extremum graphs for the input vibrational displacements (black to
red: negative to positive translations,
equilibrium: white).
\emph{(b)} Partial isomorphisms to the equilibrium state (colored arcs, white:
unmatched arcs).
\emph{(c)} Bond occurrence rate (black to red).
\emph{(d-h)} Bond occurrence rate for various configurations, as a function of
the vibrational displacement:
\emph{(d)} missing H-bond \todo{($\alpha_0$)},
\emph{(e)} misconnected H-bonds \todo{($[\alpha_1, \alpha_2]$)},
\emph{(f)} valid H-bonds \todo{($[\alpha_3, \alpha_{12}]$)},
\emph{(g)} misconnected H-bonds \todo{($[\alpha_{13}, \alpha_{17}]$)},
\emph{(h)} missing H-bonds \todo{($[\alpha_{18}, \alpha_{20}]$)}.
Valid H-bonds occur around the equilibrium,
while
misconnected and missing H-bonds
\revision{(arrows)}
occur for intermediate and
large displacements.
}
\label{fig_prismMode0}
\end{figure*}

\section{Analysis of Ensembles of Extremum Graphs}
\label{sec_ensemble}

This section describes our approach for estimating a geometry-aware partial
isomorphism between two extremum graphs, representing two distinct electron
densities. This partial isomorphism is
\revision{later}
exploited in the definition of our
stability measure (called \emph{bond occurrence rate}).


\subsection{Overview}
Let $\rho'_i : \domain \rightarrow \range$ and $\rho'_j : \domain \rightarrow
\range$ be two opposite electron densities representing a given molecular
system under distinct geometrical configurations (e.g., distinct vibrational
displacements).
Moreover, let $\extremumGraph(\rho'_i)$ and $\extremumGraph(\rho'_j)$ be their
respective extremum graphs, with their
 node sets $\minimumSet(\rho'_i)$ and $\minimumSet(\rho'_j)$ and their arc sets
 $\unstableSets(\rho'_i)$ and $\unstableSets(\rho'_j)$,
see
\autoref{sec_topological_features}.

The goal of the approach documented in this section is to establish a reliable
correspondence between these two extremum graphs. For this, we first establish
a correspondence
$\phi_{\minimumSet_{i \rightarrow j}} : \minimumSet(\rho'_i) \rightarrow
\minimumSet(\rho'_j)$
between the nodes of the extremum graphs (see \autoref{sec_atom_assignment}
and \autoref{fig_partialIsomorphism}(a)),
prior to extending it to their arcs,
in the form of an induced partial isomorphism
$\phi_{\unstableSets_{i \rightarrow j}} : \unstableSets(\rho'_i) \rightarrow
\unstableSets(\rho'_j)$ (see \autoref{sec_induced_partial_isomorphism} and
\autoref{fig_partialIsomorphism}(b)).
To do so,
we specifically exploit the fact that our database models water hexamers.
Therefore, at this stage, each extremum graph counts exactly $18$ nodes
(representing the $3$ atoms of the $6$ water molecules) \revision{and} we have
$|\minimumSet(\rho'_i)| = |\minimumSet(\rho'_j)|$, which is a valuable
simplifying hypothesis.

\subsection{Optimal node assignment}
\label{sec_atom_assignment}

While the original molecule specifications  (i.e., the
atom locations) are
known for both opposite densities $\rho'_i$ and $\rho'_j$ (this is the input of
the DFT computation,
\autoref{sec_chemistry}), in practice, the locations of their resulting
\emph{minima}
do not exactly coincide to this specification
due to both spatial quantization artifacts
and the imbalance between the atomic masses of the oxygen and hydrogen atoms
(which displaces the minima, especially those associated to hydrogen atoms).
\revision{Thus,}
a robust
algorithm needs to be considered to establish a reliable correspodence between
the node sets $\minimumSet(\rho'_i)$ and $\minimumSet(\rho'_j)$.
For this, we consider the following balanced assignment
problem
(the simplest form
of optimal transport \cite{PeyreC19}).


Let $\phi_{i \rightarrow j} :\minimumSet(\rho'_i) \rightarrow
\minimumSet(\rho'_j)$ be a bijection between the node sets
$\minimumSet(\rho'_i)$ and
$\minimumSet(\rho'_j)$.
\revision{Next,}
let
$E(\phi_{i \rightarrow j})$ be the following energy
term:
\myspace
\begin{eqnarray}
\label{eq_energy}
E(\phi_{i \rightarrow j}) = \sum_{n \in \minimumSet(\rho'_i)} || n - \phi_{i
\rightarrow j}(n) ||_2.
\end{eqnarray}
\myspace
The term
$E(\phi_{i \rightarrow j})$ simply evaluates the
sum of the distances in
3D space between the minima of $\rho'_i$
and their images by $\phi_{i \rightarrow j}$
(\revision{intuitively,}
the sum of the geometrical
\emph{mismatches} between minima).
\revision{Finally,}
we select as
correspondence
$\phi_{\minimumSet_{i \rightarrow j}}$ a global minimizer of $E(\phi_{i
\rightarrow j})$:
\myspace
\begin{eqnarray}
 \nonumber
 \phi_{\minimumSet_{i \rightarrow j}} = \argmin_{\phi_{i \rightarrow j} \in
\Phi_{i \rightarrow j}}
E(\phi_{i \rightarrow j}),
\end{eqnarray}
\myspace
where $\Phi_{i \rightarrow j}$ denotes the set of all bijections between
$\minimumSet(\rho'_i)$ and $\minimumSet(\rho'_j)$. In practice, the solution of
this assignment problem,
\autoref{fig_partialIsomorphism}(a),
can be easily obtained with either exact
\cite{Munkres1957} or approximate \cite{Bertsekas81} algorithms.


%
%
%

\subsection{Induced partial isomorphism}
\label{sec_induced_partial_isomorphism}

The optimal node assignment computed in the previous section is extended into
an induced partial isomorphism $\phi_{\unstableSets_{i \rightarrow j}}$ as
follows.
Let $a_i \in \unstableSets(\rho'_i)$ and $a_j \in \unstableSets(\rho'_j)$ be
two arcs of $\extremumGraph(\rho'_i)$ and $\extremumGraph(\rho'_j)$
respectively.
\revision{Next,}
we assign $a_i$ to $a_j$,
i.e., $\phi_{\unstableSets_{i \rightarrow j}}(a_i) = a_j$,
if and only if:
\myspace
\begin{eqnarray}
\nonumber
  \phi_{\unstableSets_{i \rightarrow j}}(a_i) =
a_j
  \Longleftrightarrow
  \begin{cases}
    \phi_{\minimumSet_{i \rightarrow j}}(a_i^1) = a_j^k\\
    \phi_{\minimumSet_{i \rightarrow j}}(a_i^2) = a_j^l
  \end{cases},
\end{eqnarray}
\myspace
where $a_i^1$ stands for the first node of $a_i$ and $a_i^2$ for the second, and
where $(k, l) \in \{(1, 2), (2, 1)\}$.
In other words, an arc $a_i$ will be assigned to an arc $a_j$ if the nodes of
$a_i$ are mapped through $\phi_{\minimumSet_{i \rightarrow j}}$ to those of
$a_j$.

If the resulting map $\phi_{\unstableSets_{i \rightarrow j}} :
\unstableSets(\rho'_i) \rightarrow \unstableSets(\rho'_j)$ is a bijection, we
will say that it describes a \emph{complete} isomorphism between
$\extremumGraph(\rho'_i)$ and $\extremumGraph(\rho'_j)$. Otherwise, it only
describes a \emph{partial} isomorphism, e.g.,
\autoref{fig_partialIsomorphism}(b).


%

\begin{figure}
\centering
%
%

\includegraphics[width=\linewidth]{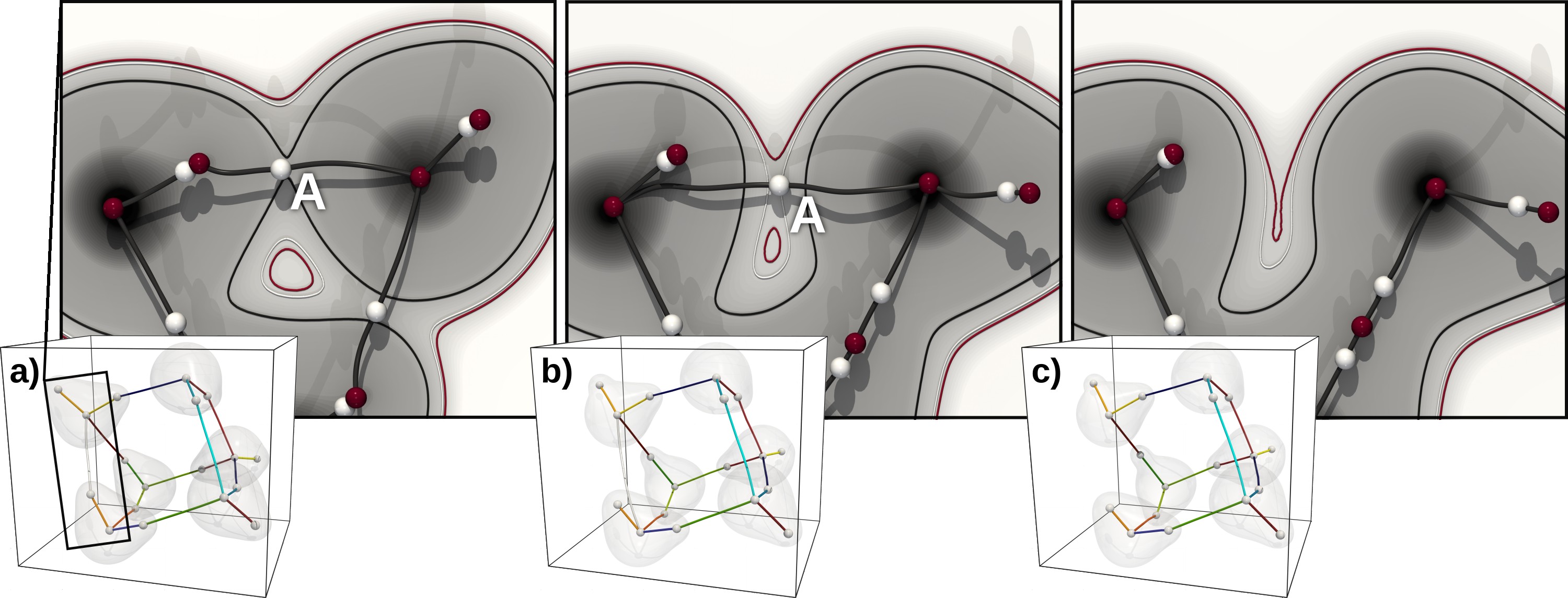}

\caption{H-bond path configurations for the \emph{Prism} (vibration mode $0$),
according to their unstable
sets (white: saddles, red: minima, black curves: unstable sets). A 2D
\revision{slice}
of
the electron density is shown in the background with three isolines. The
corresponding partial isomorphisms are shown with colors in the insets.
\emph{(a)} At equilibrium (\todo{$\alpha_{10}$}), the unstable set of the
saddle $A$
connects it
to hydrogen (left)
and oxygen (right) atoms.
\emph{(b)} Under intermediate
displacement (\todo{$\alpha_2$}), the
saddle
$A$ increases its $\rho'$ value (white isoline). The geometry of its unstable
set is altered such that it now connects it to two oxygen atoms. We
\revision{refer} to
this configuration as a \emph{misconnected H-bond}.
\emph{(c)} Under large
displacement (\todo{$\alpha_0$}), the
saddle $A$ disappears. The electron density field has been sufficiently modified
by the vibrational displacement
to cancel the corresponding saddle-saddle persistence pair
\cite{CohenSteinerEH05}.
We
\revision{refer} to this
configuration as a \emph{missing H-bond}.
}
\label{fig_bondClassification}
\end{figure}

\subsection{Bond occurrence rate}
\label{sec_stability_measure}

In this section, we leverage the
partial isomorphism
introduced above
to provide an estimation of the stability of an H-bond path.

Let $\{\extremumGraph(\rho'_1), \extremumGraph(\rho'_2), \dots,
\extremumGraph(\rho'_n)\}$ be the set of $n$ extremum graphs generated with the
approach described in \autoref{sec_topological_features}, from an ensemble of
$n$ electron densities (generated as documented in \autoref{sec_chemistry}).

Given an arc $a$ from an extremum graph $\extremumGraph(\rho'_i)$ (with $i \in
[1, n]$), we estimate its \emph{bond occurrence rate} $\occurrence(a)$ as
follows:
\myspace
\begin{eqnarray}
\nonumber
 \occurrence(a) = {{1}\over{n}} \sum_{j \in [1, n]} |\{\phi_{\unstableSets_{i
\rightarrow j}}(a) ~ | ~
~ \phi_{\unstableSets_{i
\rightarrow j}}(a) \neq \emptyset \}|.
\end{eqnarray}
\myspace
In other words, the bond occurrence rate $\occurrence(a)$ measures the number
of times a bond $a$ from an extremum graph $\extremumGraph(\rho'_i)$ has been
matched to another bond in the ensemble (divided by
the size of the ensemble).
This quantity, \autoref{fig_partialIsomorphism}(c), is expected to equal $1$
for stable bonds (e.g., covalent
bonds)
and to be in the range $[{{1}\over{n}}, 1]$
for unstable bonds.

\section{Results}
\label{sec_results}

This section presents experimental results obtained with
our approach, implemented \revision{in C++ within}
TTK
\cite{ttk17, ttk19}.
\revision{Our experimental data
\autoref{sec_chemistry} is available at this address:
%
%
%
%
\href{https://github.com/thom-dani/BondMatcher}{https://github.com/thom-dani/BondMatcher}.}
\revision{In both case studies (Secs. \ref{sec_useCase_pathways} and
\ref{sec_useCase_vibrations}), our \emph{bond occurrence rate}
(\autoref{sec_stability_measure}) played an instrumental visual role as it
enabled, with only one visualization (at equilibrium), the quick identification
of the H-bonds involved in transformations (instead of exploring one by one the
members
of the considered ensemble).
This quick identification eased their
selection for refined inspections, detailed in Secs. \ref{sec_useCase_pathways}
and
\ref{sec_useCase_vibrations}.}


\subsection{Geometrical interpretation}
\label{sec_discussion}

Running the topological data analysis pipeline described in
\autoref{sec_topology} yields a database of extremum graphs.
\autoref{fig_bondClassification} presents a classification of the H-bond
configurations that we have observed in this database, according to their
associated unstable sets (\autoref{sec_background_topology}).
This figure shows that, under vibrational
displacement, the \emph{bond critical point} (\autoref{sec_background_qtaim})
associated to an H-bond path \emph{gradually} increases its $\rho'$ value.
Depending on the amplitude of the vibrational displacement, this can result
in two degenerate configurations.
First, a \emph{misconnected H-bond}, \autoref{fig_bondClassification}(b),
occurs when the $\rho'$ value of the bond critical point increased to an
intermediate level, such that
the geometry of its unstable set is altered to the point that it
connects two oxygen atoms
together (a configuration which is not chemically relevant
in this context). Second, a \emph{missing H-bond},
\autoref{fig_bondClassification}(c), occurs when the $\rho'$ value of the bond
critical point increased so much that the corresponding saddle-saddle pair has
been canceled \cite{CohenSteinerEH05}. This figure suggests that,
\revision{depending on the
geometric transformation
induced by the vibration,}
the
transition from
a valid H-bond path at
equilibrium
to a missing one
is \emph{gradual} (via a misconnection step),
which may relate to
experimental studies
discussing the role of oxygen-oxygen contractions 
on vibrations of liquid water \cite{yang.etal_n_2021}.

\autoref{fig_prismMode0} confirms this observation by inspecting the entire set
of $21$ vibrational displacements for the fist vibrational mode of the
\emph{Prism}.
Specifically, it shows that missing H-bonds occur at the extremities of the
displacement spectrum \todo{($\alpha_0$ and $[\alpha_{18},
\alpha_{20}]$)} and that misconnected H-bonds occur within
immediately adjacent intervals \todo{($[\alpha_1, \alpha_3]$ and
$[\alpha_{13}, \alpha_{17}]$)}. This confirms that the transition from valid
(around the equilibrium),
to misconnected, to missing H-bond
is
indeed
\emph{gradual} along
\revision{this}
vibration. \autoref{fig_prismMode0}(c)
also shows the bond occurrence rate (\autoref{sec_stability_measure})
for
these configurations.
It reports lower rates for
misconnected H-bonds, indicating the transitional aspect of this configuration.


In the following, we will consider that a bond identified at equilibrium
with the procedure described in \autoref{sec_topology2chemistry} is
\emph{unstable} if its bond occurrence rate (\autoref{sec_stability_measure}) is
smaller than $1$. Otherwise, it is stable.

\begin{figure}
\centering

\includegraphics[width=\linewidth]{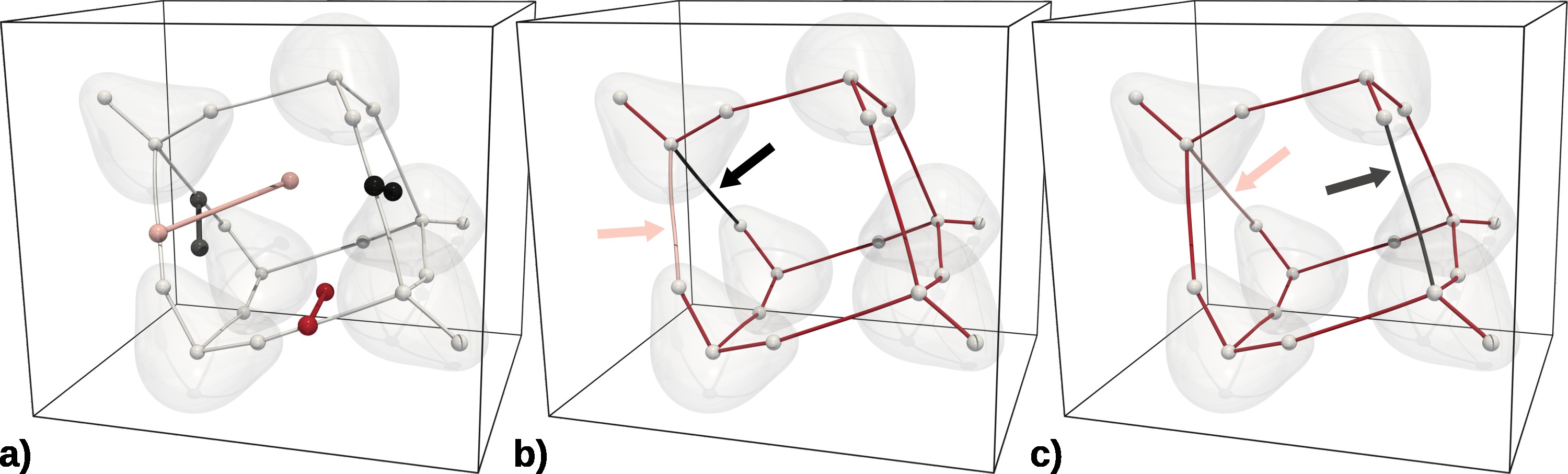}

\caption{Observing the \emph{persistence} \cite{edelsbrunner02}
and bond occurrence rate on the \emph{Prism} (equilibrium).
\emph{(a)} Saddle
persistence pairs (persistence: black
to red)
\revision{: $1$-saddles create \emph{cycles} in the extremum graph, while
$2$-saddles represent the topological discs which fill these cycles.}
\emph{(b-c)} Bond occurrence rate (black to red), for mode $0$ \emph{(b)} and
mode $2$ \emph{(c)}.
In this example, the
saddles defining unstable H-bonds
(arrows)
are also
involved in the least persistent saddle
pairs (black, gray, pink).}
\label{fig_persistenceVSstability}
\end{figure}

\autoref{fig_persistenceVSstability} investigates the relation between the
unstable bonds of the \emph{Prism} (relative to vibrations)
and
\emph{topological persistence} \cite{edelsbrunner02},
a
generic
importance measure
often used to estimate prominence and stability in Topological Data Analysis
\cite{edelsbrunner09}. In particular, this figure shows that, in this example,
the bonds identified as \emph{unstable} are defined relatively to saddles which
are also involved in the least persistent saddle-saddle pairs (similar
observations, not reported here, were made for the \emph{Cage} isomer). This
indicates that such low-persistence saddles are more likely to undergo, under
vibrations, the transitions observed in \autoref{fig_bondClassification}.
\revision{Still, note}
that in theory, instability can still  be observed with unstable sets
defined relative to the most persistent saddles, as
illustrated in
\autoref{fig_unstability}. Then, although persistence and bond stability appear
to be correlated in our experiments, it is not necessarily the case in
generality.

Finally, we conclude this section by observing that in all our experiments
(either on vibrational effects or proton tunneling), the bonds known as
\emph{covalent} have always been reported with a bond occurrence rate of $1$.
This indicates that our stability measure (\autoref{sec_stability_measure}),
for these bonds already, is concordant with established chemical expectations.


%
%
%
%

\begin{figure*}
\centering
\includegraphics[width=\linewidth]{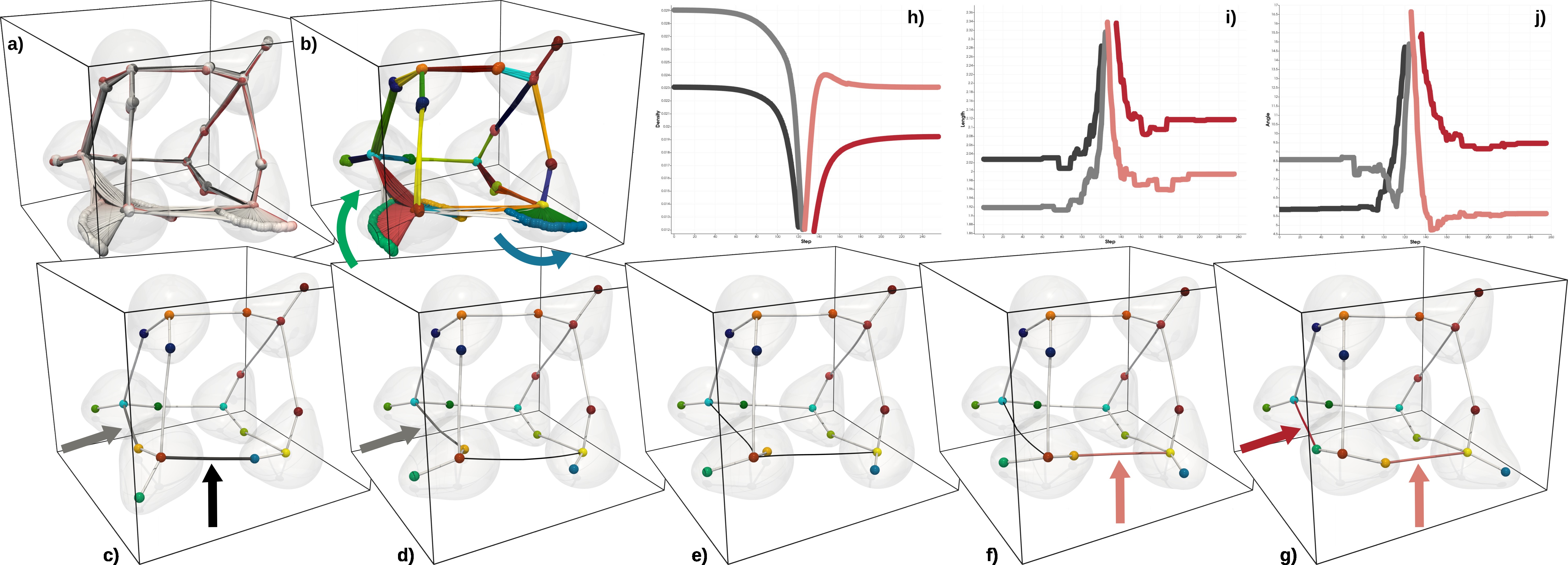}
\caption{H-Bond analysis along the geared tunneling pathway on the
\emph{Prism}.
\emph{(a)} Extremum graphs for the input pathway (time: black to red).
\emph{(b)} Partial isomorphisms to the pathway mid-point (colored arcs, white:
unmatched arcs).
\emph{(c)} From $t=0$ to $t=120$, the system
includes, among others,
two H-bonds (black and gray)  respectively involving a blue and orange
hydrogen atoms.
\emph{(d)} From $t=121$ to $t=124$, the black H-bond has disappeared
and is replaced by a misconnected H-bond (thin black cylinder).
\emph{(e)} At $t=125$, the dark gray H-bond has disappeared and
is also replaced by a misconnected H-bond.
\emph{(f)} From $t=126$ to $t=133$, a new H-bond has appeared (pink), involving
the former, orange hydrogen atom.
\emph{(g)} As of $t=134$ (till $255$), a new H-bond has appeared (red),
involving a green hydrogen atom.
\emph{(h-j)} Bond critical point density,
(Ac$-$H)
bond
length and
(Ac$-$Dn, Ac$-$H)
bond angle over time for the black, gray, pink and red H-bonds.
Our analysis captures the structural change involved in this pathway, with a
clear \emph{breaking point} for the two H-bonds involved in this geared
tunneling motion, characterized by a sharp decrease in bond critical point
densities (below $1.5 \times 10^{-2} $), as well as a sudden increase in bond
length (above $2.3$ \r{A}) and
angle (above $14$°).
}
\label{fig_antigeared}
\end{figure*}

\begin{figure}
\centering
\includegraphics[width=\linewidth]{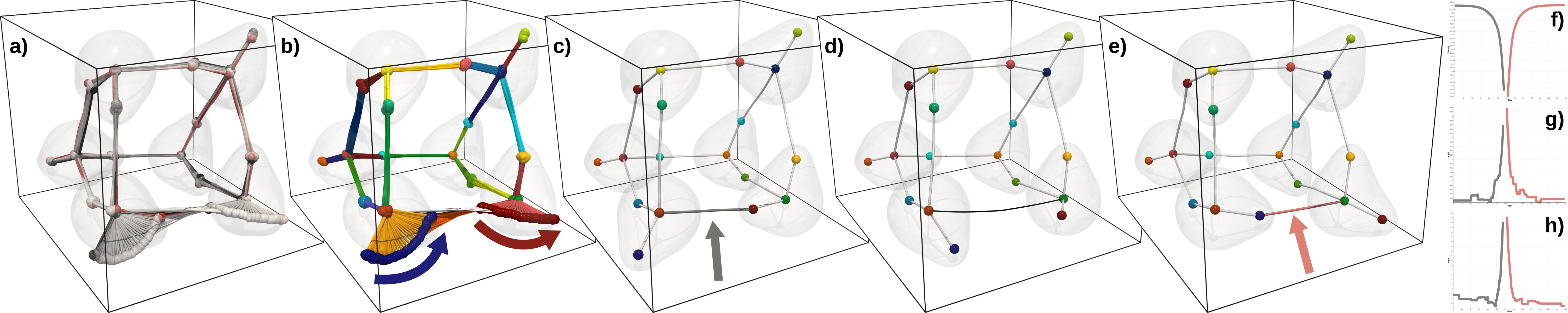}
\caption{H-Bond analysis along the anti-geared tunneling pathway on the
\emph{Prism}.
\emph{(a)} Extremum graphs for the input pathway (time: black to red).
\emph{(b)} Partial isomorphisms to the pathway mid-point (colored arcs, white:
unmatched arcs).
\emph{(c)} From $t=0$ to $t=115$, the system
includes
an H-bond (gray)
involving
a red
hydrogen atom.
\emph{(d)} From $t=116$ to $t=123$, the gray H-bond has disappeared and
is replaced by a misconnected H-bond (thin black cylinder).
\emph{(e)} As of $t=124$ (till $255$), a new H-bond has appeared (pink),
involving a blue hydrogen atom.
\emph{(f-h)} Bond critical point density,
(Ac$-$H)
bond
length and
(Ac$-$Dn, Ac$-$H)
bond
angle over time for the gray and pink H-bonds.
Our analysis captures the structural change involved in this pathway, with a
clear \emph{breaking point}, similar to the geared motion:
%
%
bond critical point
density below $1.5 \times 10^{-2} $,  sudden increase in bond
length (above $2.3$ \r{A}) and
angle (above $14$°).
}
\label{fig_geared}
\end{figure}

\subsection{Case Study 1: H-Bond stability in tunneling reaction}
\label{sec_useCase_pathways}
To evaluate the chemical relevance of our work, we first study the
\emph{Prism} under
\emph{proton tunneling} (\autoref{sec_chemistry_pathways}), a
phenomenon for
which
theoretical models have been validated by experimental
observations \cite{rotations16}.

\autoref{fig_antigeared} presents the results obtained by our approach,
for the pathway describing a \emph{geared} tunneling
motion, where two water molecules of the \emph{Prism} rotate
simultaneously, resulting in a characteristic switch in the configuration of
H-bonds \cite{rotations16}. This figure reports the geometry-aware partial
isomorphisms to the pathway mid-point, as computed by our approach
(\autoref{sec_ensemble}). Specifically, the bond matching provided by our
method enables an individual analysis of the geometrical properties of each
bond through time. In particular, this analysis identifies two
H-bonds with an occurrence rate lower than $1$
(black and gray), which successively disappear (at $t=121$, then $t=124$).
Symmetrically, two new H-bonds (pink and red) successively appear (at $t=126$
and $t=134$). As shown on the companion plots, the disappearance of the black
and gray H-bonds is associated with a drastic decrease in the respective bond
critical point density ($\rho$), along with an abrupt increase in
(Ac$-$H)
bond length
and
(Ac$-$Dn, Ac$-$H)
bond angle.
The resulting spikes in the
plots materialize a \emph{breaking point} in these geometrical indicators,
beyond which H-bond paths disappear. A symmetrical observation can be made for
the
appearing H-bonds (pink and red). These plots, enabled by our approach, confirm
the brevity of the H-bond configuration switch ($14$ time steps out of $255$).
They
also provide limit values, for the bond critical point density ($1.5
\times 10^{-2}$), bond length ($2.3$ \r{A}) and angle ($14$°), beyond which an
H-bond path can be considered as disappearing.
\autoref{fig_geared} provides a similar analysis,
with identical conclusions,
for
an \emph{anti-geared} tunneling motion, where only one H-bond
disappears (and re-appears) along the pathway.

\subsection{Case-Study 2: H-Bond stability in molecular vibrations}
\label{sec_useCase_vibrations}
\begin{figure}
\includegraphics[width=0.475\linewidth]{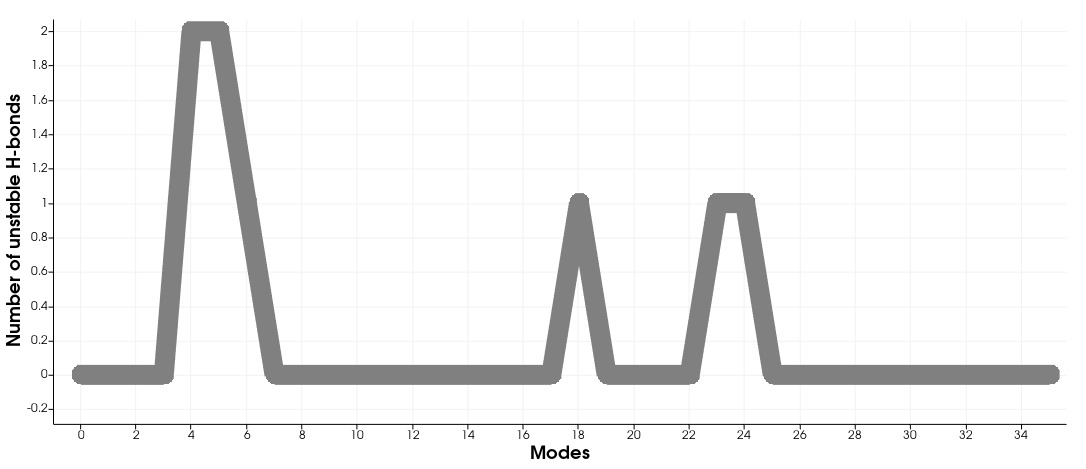}
\hfill
\includegraphics[width=0.475\linewidth]{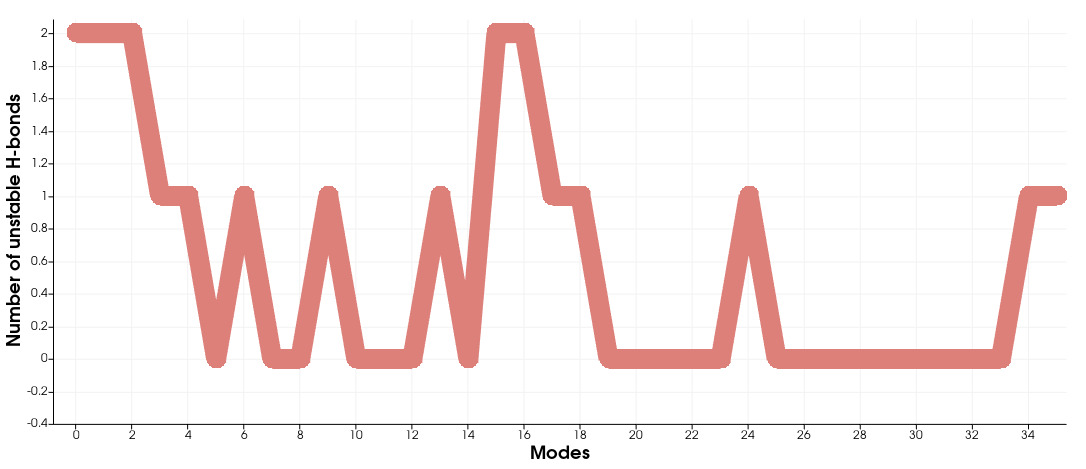}

\caption{Number of unstable H-bond paths as a function of the vibration mode
number,
for the \emph{Cage} (left) and the \emph{Prism} (right). No unstable H-bond
paths
are reported by our approach for the \emph{Ring} and the \emph{Book}.}
\label{fig_unstableModes}
\end{figure}

\begin{figure}
\centering
%
%
%
\includegraphics[width=\linewidth]{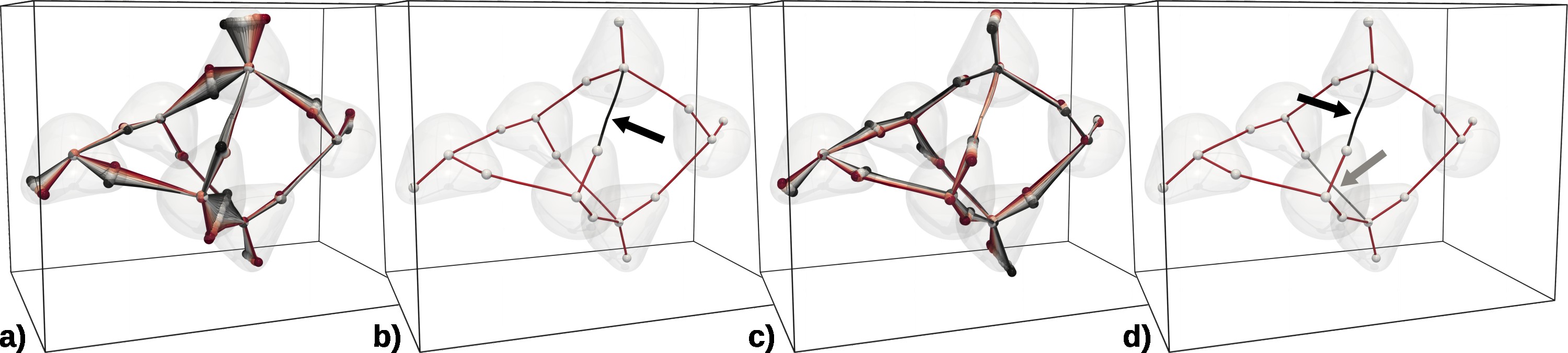}
\caption{Bond occurrence rate  for the \emph{Cage} at two vibration
modes (left: mode $18$, right: mode $5$).
\emph{(a,c)} Extremum graphs for the input vibrational displacements (black to
red).
\emph{(b)}
At mode $18$, only one H-bond path is reported as
unstable (black).
\emph{(d)}
At mode $5$, two H-bond paths are unstable
(black and gray).
At mode $18$,
the breaking of the unstable H-bond
\emph{(b)} makes the extremum graph become isomorphic to
the \emph{Book}.}
\label{fig:vib_cage2book}
\end{figure}

We now investigate the stability of H-bond paths under molecular vibrations,
with
the data generated as specified in \autoref{sec_chemistry_vibrations}.
For this, for each considered isomer of water hexamer (the \emph{Ring}, the
\emph{Book}, the \emph{Cage} and the \emph{Prism},
\autoref{sec_chemistry_configuration_selection}), we analyzed $21$ vibrational
displacements under $48$ vibrational modes.
\autoref{fig_unstableModes} reports the number of H-bonds identified at
equilibrium (\autoref{sec_topology2chemistry}), which have been reported as
\emph{unstable} (i.e., with a bond occurrence rate smaller than $1$,
\autoref{sec_stability_measure}), as a function of the vibration mode. The
first important observation is that the \emph{Ring} and the \emph{Book} do
not report any unstable bonds.
In contrast, the \emph{Cage} reports few vibration modes for
which unstable H-bonds are discovered, while the \emph{Prism} reports more.
Note that, for the latter two isomers, the analysis had to be restricted to the
modes $0$ to $36$, as beyond, for each mode, several electron density datasets
were lacking some hydrogen atoms (i.e., they were initially exhibiting less than
$18$ minima of $\rho'$, see \autoref{sec_limitations}). Specifically, the
\emph{Cage} reports $2$ H-bond paths at equilibrium, which have been identified
as
unstable in at least one vibration mode (\autoref{fig:vib_cage2book}).
The \emph{Prism} reports $3$ unstable H-bond paths
(visible in \autoref{fig_persistenceVSstability}).

\begin{figure}
\centering
%

\includegraphics[width=\linewidth]{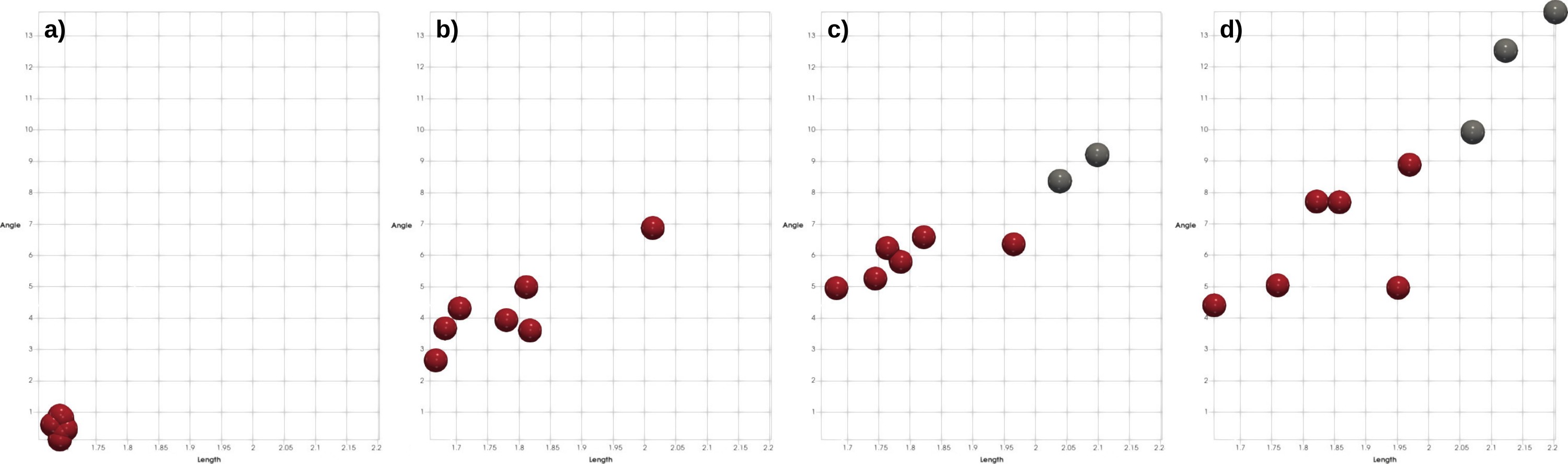}

\caption{Geometrical indicators,
(Ac$-$Dn, Ac$-$H) bond angle
as a function of
(Ac$-$H)
bond length,
for the H-bonds identified at equilibrium for the \emph{(a) Ring},
\emph{(b) Book}, \emph{(c) Cage}, and \emph{(d) Prism}. The H-bonds
reported as unstable in at least one vibration mode are shown in gray
(red otherwise).}
\label{fig_unstableDescriptors}
%
\end{figure}

\autoref{fig_unstableDescriptors} investigates the geometrical characteristics,
at equilibrium, of the different identified H-bonds
(\autoref{sec_topology2chemistry}) for the $4$ isomers considered in our study.
In particular, in these plots, each H-bond is represented by a point: $6$ for
the \emph{Ring}, $7$ for the \emph{Book}, $8$ for the \emph{Cage} and $9$ for
the \emph{Prism}. Moreover, the H-bonds reported as \emph{unstable} (with an
occurrence rate below $1$) are marked in gray. The purpose of this experiment
is to observe if the unstability of an H-bond under molecular vibrations can be
predicted based on its geometry at equilibrium, in particular in the light of
the limit values identified in the previous case study
(\autoref{sec_useCase_pathways}).

\autoref{fig_unstableDescriptors} confirms that, at equilibrium, the
geometrical
indicators
(bond length
and angle) of the H-bonds of the \emph{Ring} and \emph{Book} are associated to
 different ranges than for the \emph{Cage} and \emph{Prism}.
This is obvious with the \emph{Ring}, which has a highly
symmetrical structure (\autoref{fig:teaser}), leading to compact point
distributions, \autoref{fig_unstableDescriptors}(a). Also, these values are
far from the limit values identified in \autoref{sec_useCase_pathways}. In
contrast, the H-bonds identified as unstable in the \emph{Cage} and
\emph{Prism} (gray points) are also the H-bonds whose geometrical indicators
are the closest to the limit values identified in
\autoref{sec_useCase_pathways}.
This
indicates that these unstable H-bonds are already, initially at equilibrium,
close
to the H-bond breaking points identified in the tunneling case
study (\autoref{sec_useCase_pathways}), and that they are, consequently, more
conducive to disappearance under molecular vibrations.
As discussed in
\autoref{fig:vib_cage2book}, such an H-bond disappearance can eventually
transform an isomer into another.
Then, the range of
geometrical indicators for unstable H-bonds (gray spheres) -- bond length: $(2,
2.3)$ \r{A}, bond angle: $(8, 14)$° -- can be interpreted as uncertainty
intervals, where H-bond paths are unstable and likely to disappear under
vibrations.

\subsection{Computational aspects}
\label{sec_computationalAspects}

The timings for the quantum chemistry calculations
are approximately $6000$ seconds for the geometry optimization 
and frequency calculations for each isomer 
(\autoref{sec_useCase_vibrations})
and $300$ seconds for each (single-point) electron density 
computation.
Our
topological analysis pipeline (\autoref{sec_topology}) involves
algorithms whose running time is at most quadratic, in the worst case, with the
number of vertices in the
grid. In practice,
this pipeline required approximately $200$ seconds of computation on a
desktop computer for each electron density.
Once the database of electron densities is
processed, a database of extremum graphs is available. Its processing by the
approach described in \autoref{sec_ensemble} was very fast in practice, as each
extremum graph is extremely small ($18$ nodes), resulting in runtimes below a
second for each ensemble, even when considering the largest ensembles ($256$
members for each tunneling pathway).

\ifdefined\includeSuggestions
\color{purple}

\color{black}
\fi

\subsection{Limitations}
\label{sec_limitations}

A first limitation, which prevented us from analyzing vibrations modes beyond
mode $36$ for the \emph{Cage} and \emph{Prism}, is that our input database
contains a few datasets lacking some critical points corresponding to
hydrogen atoms, with a number of local
minima of $\rho'$ initially smaller than $18$. 
These are high-frequency modes describing the O$-$H 
stretching involving the non-H-bonded H atoms.
These modes may be susceptible to effects not explicitly
handled
in the
harmonic oscillator model.
Moreover, our model artifically extends the viable range 
of these motion amplitudes.
\revision{As a result,}
the displacements of these H atoms may
be overestimated, such that their
corresponding minima
may sometimes vanish.
This might be addressed by re-generating
the data on a finer grid (e.g., $512^3$), at the expense of
larger computation times. Another possibility would be to consider an
unbalanced variant of the assignment problem
from
\autoref{sec_atom_assignment}.
\revision{Nevertheless,} this would result in certain covalent
bonds (involving the missing hydrogen atoms) being reported as unstable,
which would challenge
the overall interpretation.

In principle, large atom displacements may also challenge our optimal node
assignment procedure (\autoref{sec_atom_assignment}),
in which case,
further costs may need to be
considered to extend the assignment energy with domain knowledge.
However, we have not observed extreme displacements
of
minima in our experiments,
despite extending the range of amplitudes of atomic motions 
beyond the classical harmonic model (\autoref{sec_chemistry_vibrations}).
Our database considers intrinsic molecular vibrations
that
result
in practice
in moderate
displacements of minima, such that our node matching
procedure (\autoref{sec_atom_assignment}) always provided relevant
assignments.


%
%


\section{Conclusion}
\label{sec_conclusion}

This
application
paper presented \emph{BondMatcher}, a framework for studying the stability
of QTAIM H-bond paths in molecular systems.
Specifically, by computing
geometry-aware
partial isomorphisms between extremum
graphs extracted with discrete Morse theory (DMT), our work enables a robust
tracking of H-bond paths in ensembles of electron densities. This
detailed
per-bond tracking enabled the identification of unstable H-bond paths,
as well as their individual geometrical analysis through the ensemble.
%
This
led to the following empirical
findings:
\begin{itemize}
 \item Despite their theoretical instability
 (\autoref{fig_unstability}),
QTAIM bond paths
have shown in our experiments to be
surprisingly stable under the \emph{vibrational} perturbations
typically studied in chemistry (with only $2$ and $3$ H-bonds identified as
unstable for the \emph{Cage} and \emph{Prism}).
\item The saddles defining
H-bond paths
that are
unstable under
vibrations are also the saddles
associated to the lowest \emph{persistence}
(\autoref{fig_persistenceVSstability}).
This
indicates that
low-persistence saddle pairs are likely to be canceled
first by vibrations.
\revision{Still}
in principle, path instability can also be
observed 
for persistent saddles
(see \autoref{fig_unstability}).
\item Our experiments (\autoref{fig_prismMode0}) suggest that the
disappearance of an H-bond path under vibrations
is
\emph{gradual}, involving
first a \emph{mis-connection} (that may be related to documented oxygen-oxygen
contractions \cite{yang.etal_n_2021}), followed by a cancellation of its
saddle.
\item Our experiments on documented proton tunneling (Figs.
\ref{fig_antigeared}, \ref{fig_geared}) confirm the brevity of H-bond
reconfiguration under this mechanism
\cite{rotations16}. Moreover, this analysis
identifies clear \emph{breaking points} in terms of (Ac$-$H) bond length
($2.3$ \r{A}) and (Ac$-$Dn, Ac$-$H) bond angle ($14$°), beyond which a
H-bond path disappears.
\item Finally, the geometrical analysis, at equilibrium, of the H-bond paths
identified as unstable under vibrations (\autoref{fig_unstableDescriptors})
reveals a correlation between our stability measure (bond occurrence rate,
\autoref{sec_stability_measure}) and the above geometrical indicators. In
particular, the most unstable H-bond paths are also the bonds whose indicators
are the closest to the above limit values defining a breaking point. Then, the
indicator ranges for unstable H-bond paths -- bond length: ($2$, $2.3$) \r{A},
bond angle: ($8$, $14$)° -- can be interpreted as uncertainty intervals, within
which H-bond paths are unstable and are likely to disappear under molecular
vibrations (indicating modes that can contribute to isomerization of (H$_2$O)$_6$,
\autoref{fig:vib_cage2book}).
\end{itemize}

We believe our work opens several research avenues at the intersection between
DMT and QTAIM. First, our \emph{BondMatcher} framework can be
used for other studies requiring a robust, per-bond comparative analysis of
several systems. For instance, the protocol used in our case studies could be
extended to more complex
molecular systems
\revision{(e.g., as in \cite{harshChemistry})}, to get a
refined picture of the instability of H-bond paths.
Second, our framework could also be used for the comparative analysis of a
single molecular system, but studied under various
descriptors (e.g.,
electron density versus bare nuclear potential). Such feature correspondences
between descriptors could be instrumental to accelerate several high-throughput 
chemical
analysis tasks, as the electron density can still be computationally expensive to
obtain. However, such comparisons would require to revisit our extremum graph
matching procedure, in particular to reliably account for distinct extremum
counts.

\ifdefined\includeSuggestions
\color{purple}

\color{black}
\fi



\clearpage

\acknowledgments{%
\scriptsize{%
This work is partially supported by the European Commission
grant ERC-2019-COG \emph{``TORI''} (ref. 863464,
\url{https://erc-tori.github.io/})
\gosia{and the Polish National Science Centre (NCN) (grant 2020/38/E/ST4/00614).
We gratefully acknowledge the Poland's high-performance Infrastructure PLGrid 
(HPC Center: ACK Cyfronet AGH) for providing computer facilities and support 
within computational grant no PLG/2024/017638.
We also thank Prof. Dr. Jeremy Richardson (ETH Z\"urich, Switzerland)
for providing molecular structures of the (H$_2$O)$_6$ prism isomer modeling 
the tunneling reactions.
}
}}

\bibliographystyle{abbrv-doi-hyperref}

\bibliography{templateCLEANED}

\begin{thebibliography}{10}

\bibitem{arunan.etal_pac_2011}
E.~Arunan, G.~R. Desiraju, R.~A. Klein, J.~Sadlej, S.~Scheiner, I.~Alkorta,
  D.~C. Clary, R.~H. Crabtree, J.~J. Dannenberg, P.~Hobza, H.~G. Kjaergaard,
  A.~C. Legon, B.~Mennucci, and D.~J. Nesbitt.
\newblock Definition of the hydrogen bond ({IUPAC} {Recommendations} 2011).
\newblock {\em Pure and Applied Chemistry}, 83(8):1637--1641, 2011.
  \href{https://doi.org/10.1351/PAC-REC-10-01-02}
{doi: {{%
10\hspace{.1pt}\discretionary{.}{%
}{.}\hspace{.4pt}1351\discretionary{/}{%
}{/}PAC\discretionary{%
}{-}{-}REC\discretionary{%
}{-}{-}10\discretionary{%
}{-}{-}01\discretionary{%
}{-}{-}02}}}


\bibitem{athawale_tvcg19}
T.~M. Athawale, D.~Maljovec, L.~Yan, C.~R. Johnson, V.~Pascucci, and B.~Wang.
\newblock Uncertainty visualization of 2{D} morse complex ensembles using
  statistical summary maps.
\newblock {\em IEEE TVCG}, 28(4):1955--1966, 2022.
  \href{https://doi.org/10.1109/TVCG.2020.3022359}
{doi: {{%
10\hspace{.1pt}\discretionary{.}{%
}{.}\hspace{.4pt}1109\discretionary{/}{%
}{/}TVCG\hspace{.1pt}\discretionary{.}{%
}{.}\hspace{.4pt}2020\hspace{.1pt}\discretionary{.}{%
}{.}\hspace{.4pt}3022359}}}


\bibitem{bader94}
R.~Bader.
\newblock {\em Atoms in Molecules: A Quantum Theory}.
\newblock 1994. \href{https://doi.org/10.1093/oso/9780198551683.001.0001}
{doi: {{%
10\hspace{.1pt}\discretionary{.}{%
}{.}\hspace{.4pt}1093\discretionary{/}{%
}{/}oso\discretionary{/}{%
}{/}9780198551683\hspace{.1pt}\discretionary{.}{%
}{.}\hspace{.4pt}001\hspace{.1pt}\discretionary{.}{%
}{.}\hspace{.4pt}0001}}}


\bibitem{Bertsekas81}
D.~P. Bertsekas.
\newblock A new algorithm for the assignment problem.
\newblock {\em Math. Programming}, 21(1):152--171, 1981.
  \href{https://doi.org/10.1007/BF01584237}
{doi: {{%
10\hspace{.1pt}\discretionary{.}{%
}{.}\hspace{.4pt}1007\discretionary{/}{%
}{/}BF01584237}}}


\bibitem{harshChemistry}
H.~Bhatia, A.~G. Gyulassy, V.~Lordi, J.~E. Pask, V.~Pascucci, and P.-T. Bremer.
\newblock {TopoMS}: Comprehensive topological exploration for molecular and
  condensed-matter systems.
\newblock {\em J. of Computational Chemistry}, 39(16):936--952, 2018.
  \href{https://doi.org/10.1002/JCC.25181}
{doi: {{%
10\hspace{.1pt}\discretionary{.}{%
}{.}\hspace{.4pt}1002\discretionary{/}{%
}{/}JCC\hspace{.1pt}\discretionary{.}{%
}{.}\hspace{.4pt}25181}}}


\bibitem{biasotti08}
S.~Biasotti, D.~Giorgio, M.~Spagnuolo, and B.~Falcidieno.
\newblock Reeb graphs for shape analysis and applications.
\newblock {\em TCS}, 392(1-3):5--22, 2008.
  \href{https://doi.org/10.1016/J.TCS.2007.10.018}
{doi: {{%
10\hspace{.1pt}\discretionary{.}{%
}{.}\hspace{.4pt}1016\discretionary{/}{%
}{/}J\hspace{.1pt}\discretionary{.}{%
}{.}\hspace{.4pt}TCS\hspace{.1pt}\discretionary{.}{%
}{.}\hspace{.4pt}2007\hspace{.1pt}\discretionary{.}{%
}{.}\hspace{.4pt}10\hspace{.1pt}\discretionary{.}{%
}{.}\hspace{.4pt}018}}}


\bibitem{ttk19}
T.~Bin~Masood, J.~Budin, M.~Falk, G.~Favelier, C.~Garth, C.~Gueunet,
  P.~Guillou, L.~Hofmann, P.~Hristov, A.~Kamakshidasan, C.~Kappe, P.~Klacansky,
  P.~Laurin, J.~Levine, J.~Lukasczyk, D.~Sakurai, M.~Soler, P.~Steneteg,
  J.~Tierny, W.~Usher, J.~Vidal, and M.~Wozniak.
\newblock {An Overview of the Topology ToolKit}.
\newblock In {\em TopoInVis}, pp. 327--342. Springer, Cham, 2019.
  \href{https://dx.doi.org/10.1007/978-3-030-83500-2_16}
{doi: {{%
10\hspace{.1pt}\discretionary{.}{%
}{.}\hspace{.4pt}1007\discretionary{/}{%
}{/}978\discretionary{%
}{-}{-}3\discretionary{%
}{-}{-}030\discretionary{%
}{-}{-}83500\discretionary{%
}{-}{-}2\_16}}}


\bibitem{topoAngler}
A.~Bock, H.~Doraiswamy, A.~Summers, and C.~T. Silva.
\newblock {TopoAngler: Interactive Topology-Based Extraction of Fishes}.
\newblock {\em IEEE TVCG}, 24(1):812--821, 2018.
  \href{https://doi.org/10.1109/TVCG.2017.2743980}
{doi: {{%
10\hspace{.1pt}\discretionary{.}{%
}{.}\hspace{.4pt}1109\discretionary{/}{%
}{/}TVCG\hspace{.1pt}\discretionary{.}{%
}{.}\hspace{.4pt}2017\hspace{.1pt}\discretionary{.}{%
}{.}\hspace{.4pt}2743980}}}


\bibitem{bremer_tvcg11}
P.~Bremer, G.~Weber, J.~Tierny, V.~Pascucci, M.~Day, and J.~Bell.
\newblock Interactive exploration and analysis of large scale simulations using
  topology-based data segmentation.
\newblock {\em IEEE TVCG}, 17(9):1307--1324, 2011.
  \href{https://doi.org/10.1109/TVCG.2010.253}
{doi: {{%
10\hspace{.1pt}\discretionary{.}{%
}{.}\hspace{.4pt}1109\discretionary{/}{%
}{/}TVCG\hspace{.1pt}\discretionary{.}{%
}{.}\hspace{.4pt}2010\hspace{.1pt}\discretionary{.}{%
}{.}\hspace{.4pt}253}}}


\bibitem{carr00}
H.~Carr, J.~Snoeyink, and U.~Axen.
\newblock Computing contour trees in all dimensions.
\newblock In {\em Symp. on Dis. Alg.}, pp. 918--926. SIAM, Philadelphia, 2000.
  \href{https://doi.org/10.1016/S0925-7721(02)00093-7}
{doi: {{%
10\hspace{.1pt}\discretionary{.}{%
}{.}\hspace{.4pt}1016\discretionary{/}{%
}{/}S0925\discretionary{%
}{-}{-}7721\discretionary{%
}{(}{(}02\discretionary{)}{%
}{)}00093\discretionary{%
}{-}{-}7}}}


\bibitem{chemistry10}
F.~Cazals, F.~Cazal, and T.~Lewiner.
\newblock Molecular shape analysis based upon the {Morse-Smale} complex and the
  {Connolly} function.
\newblock In {\em SoCG}, pp. 351--360. ACM, 01 2003.
  \href{https://doi.org/10.1145/777842.777845}
{doi: {{%
10\hspace{.1pt}\discretionary{.}{%
}{.}\hspace{.4pt}1145\discretionary{/}{%
}{/}777842\hspace{.1pt}\discretionary{.}{%
}{.}\hspace{.4pt}777845}}}


\bibitem{chemistry12}
M.~Chavent, A.~Vanel, A.~Tek, B.~Levy, S.~Robert, B.~Raffin, and M.~Baaden.
\newblock {GPU-accelerated atom and dynamic bond visualization using
  hyperballs: a unified algorithm for balls, sticks and hyperboloids}.
\newblock {\em Comp. Chem.}, 32(13):2924--2935, 2011.
  \href{https://doi.org/10.1002/jcc.21861}
{doi: {{%
10\hspace{.1pt}\discretionary{.}{%
}{.}\hspace{.4pt}1002\discretionary{/}{%
}{/}jcc\hspace{.1pt}\discretionary{.}{%
}{.}\hspace{.4pt}21861}}}


\bibitem{CohenSteinerEH05}
D.~Cohen-Steiner, H.~Edelsbrunner, and J.~Harer.
\newblock Stability of persistence diagrams.
\newblock {\em DCG}, 37:103--120, 2007.
  \href{https://doi.org/10.1007/s00454-006-1276-5}
{doi: {{%
10\hspace{.1pt}\discretionary{.}{%
}{.}\hspace{.4pt}1007\discretionary{/}{%
}{/}s00454\discretionary{%
}{-}{-}006\discretionary{%
}{-}{-}1276\discretionary{%
}{-}{-}5}}}


\bibitem{vibrations1}
W.~T.~S. Cole, J.~D. Farrell, A.~A. Sheikh, {\"O}.~Y{\"o}nder, R.~S. Fellers,
  M.~R. Viant, D.~J. Wales, and R.~J. Saykally.
\newblock Terahertz vrt spectroscopy of the water hexamer-d12 prism: Dramatic
  enhancement of bifurcation tunneling upon librational excitation.
\newblock {\em The Journal of Chemical Physics}, 148(9):094301, 2018.
  \href{https://doi.org/10.1063/1.5006195}
{doi: {{%
10\hspace{.1pt}\discretionary{.}{%
}{.}\hspace{.4pt}1063\discretionary{/}{%
}{/}1\hspace{.1pt}\discretionary{.}{%
}{.}\hspace{.4pt}5006195}}}


\bibitem{vibrations2}
W.~T.~S. Cole, Özlem Yönder, A.~A. Sheikh, R.~S. Fellers, M.~R. Viant, R.~J.
  Saykally, J.~D. Farrell, and D.~J. Wales.
\newblock Terahertz vrt spectroscopy of the water hexamer-h12 cage: Dramatic
  libration-induced enhancement of hydrogen bond tunneling dynamics.
\newblock {\em The Journal of Physical Chemistry A}, 122(37):7421--7426, 2018.
  \href{https://doi.org/10.1021/acs.jpca.8b05777}
{doi: {{%
10\hspace{.1pt}\discretionary{.}{%
}{.}\hspace{.4pt}1021\discretionary{/}{%
}{/}acs\hspace{.1pt}\discretionary{.}{%
}{.}\hspace{.4pt}jpca\hspace{.1pt}\discretionary{.}{%
}{.}\hspace{.4pt}8b05777}}}


\bibitem{chemistry15}
M.~Connolly.
\newblock Analytical molecular surface calculation.
\newblock {\em Journal of Applied Crystallography}, 16:548--558, 1983.
  \href{https://doi.org/10.1107/s0021889883010985}
{doi: {{%
10\hspace{.1pt}\discretionary{.}{%
}{.}\hspace{.4pt}1107\discretionary{/}{%
}{/}s0021889883010985}}}


\bibitem{CorreaLB11}
C.~D. Correa, P.~Lindstrom, and P.~Bremer.
\newblock Topological spines: {A} structure-preserving visual representation of
  scalar fields.
\newblock {\em IEEE TVCG}, 17(12):1842--1851, 2011.
  \href{https://doi.org/10.1109/TVCG.2011.244}
{doi: {{%
10\hspace{.1pt}\discretionary{.}{%
}{.}\hspace{.4pt}1109\discretionary{/}{%
}{/}TVCG\hspace{.1pt}\discretionary{.}{%
}{.}\hspace{.4pt}2011\hspace{.1pt}\discretionary{.}{%
}{.}\hspace{.4pt}244}}}


\bibitem{DasSN24}
S.~Das, R.~Sridharamurthy, and V.~Natarajan.
\newblock Time-varying extremum graphs.
\newblock {\em CGF}, 43, 2024. \href{https://doi.org/10.1111/cgf.15162}
{doi: {{%
10\hspace{.1pt}\discretionary{.}{%
}{.}\hspace{.4pt}1111\discretionary{/}{%
}{/}cgf\hspace{.1pt}\discretionary{.}{%
}{.}\hspace{.4pt}15162}}}


\bibitem{edelsbrunner09}
H.~Edelsbrunner and J.~Harer.
\newblock {\em Computational Topology: An Introduction}.
\newblock American Mathematical Society, 2009.

\bibitem{edelsbrunner02}
H.~Edelsbrunner, D.~Letscher, and A.~Zomorodian.
\newblock {Topological Persistence and Simplification}.
\newblock {\em DCG}, 28(4):511--533, 2002.
  \href{https://doi.org/10.1007/S00454-002-2885-2}
{doi: {{%
10\hspace{.1pt}\discretionary{.}{%
}{.}\hspace{.4pt}1007\discretionary{/}{%
}{/}S00454\discretionary{%
}{-}{-}002\discretionary{%
}{-}{-}2885\discretionary{%
}{-}{-}2}}}


\bibitem{edelsbrunner90}
H.~Edelsbrunner and E.~P. Mucke.
\newblock Simulation of simplicity: a technique to cope with degenerate cases
  in geometric algorithms.
\newblock {\em ACM Trans. on Graphics}, 9(1):66--104, 1990.
  \href{https://doi.org/10.1145/77635.77639}
{doi: {{%
10\hspace{.1pt}\discretionary{.}{%
}{.}\hspace{.4pt}1145\discretionary{/}{%
}{/}77635\hspace{.1pt}\discretionary{.}{%
}{.}\hspace{.4pt}77639}}}


\bibitem{focke.etal_jcp_2024}
K.~Focke, M.~De~Santis, M.~Wolter, J.~A. Martinez~B, V.~Vallet, A.~S.
  Pereira~Gomes, M.~Olejniczak, and C.~R. Jacob.
\newblock Interoperable workflows by exchanging grid-based data between
  quantum-chemical program packages.
\newblock {\em The Journal of Chemical Physics}, 160(16):162503, 2024.
  \href{https://doi.org/10.1063/5.0201701}
{doi: {{%
10\hspace{.1pt}\discretionary{.}{%
}{.}\hspace{.4pt}1063\discretionary{/}{%
}{/}5\hspace{.1pt}\discretionary{.}{%
}{.}\hspace{.4pt}0201701}}}


\bibitem{forman98}
R.~Forman.
\newblock {A User's Guide to Discrete Morse Theory}.
\newblock {\em AM}, 1998.

\bibitem{gao.etal_jcs_2024}
C.-Y. Gao, Y.-Y. Ma, Q.~Chen, and S.-D. Li.
\newblock Fluxional {{Hydrogen Bonds}} in {{Small Water Clusters}} ({{H2O}})n
  (n\,=\,2--6).
\newblock {\em Journal of Cluster Science}, 35(2):693--700, 2024.
  \href{https://doi.org/10.1007/s10876-023-02503-x}
{doi: {{%
10\hspace{.1pt}\discretionary{.}{%
}{.}\hspace{.4pt}1007\discretionary{/}{%
}{/}s10876\discretionary{%
}{-}{-}023\discretionary{%
}{-}{-}02503\discretionary{%
}{-}{-}x}}}


\bibitem{gao.etal_sr_2022}
Y.~Gao, H.~Fang, K.~Ni, and Y.~Feng.
\newblock Water clusters and density fluctuations in liquid water based on
  extended hierarchical clustering methods.
\newblock {\em SR}, 12(1):8036, 2022.
  \href{https://doi.org/10.1038/s41598-022-11947-6}
{doi: {{%
10\hspace{.1pt}\discretionary{.}{%
}{.}\hspace{.4pt}1038\discretionary{/}{%
}{/}s41598\discretionary{%
}{-}{-}022\discretionary{%
}{-}{-}11947\discretionary{%
}{-}{-}6}}}


\bibitem{goldstein.etal__2016}
H.~Goldstein, C.~P. Poole, and J.~L. Safko.
\newblock {\em Classical Mechanics}.
\newblock Pearson, 2016.

\bibitem{gueunet_tpds19}
C.~Gueunet, P.~Fortin, J.~Jomier, and J.~Tierny.
\newblock {Task-Based Augmented Contour Trees with Fibonacci Heaps}.
\newblock {\em {IEEE} TPDS}, 30(8):1889--1905, 2019.
  \href{https://doi.org/10.1109/TPDS.2019.2898436}
{doi: {{%
10\hspace{.1pt}\discretionary{.}{%
}{.}\hspace{.4pt}1109\discretionary{/}{%
}{/}TPDS\hspace{.1pt}\discretionary{.}{%
}{.}\hspace{.4pt}2019\hspace{.1pt}\discretionary{.}{%
}{.}\hspace{.4pt}2898436}}}


\bibitem{gueunet_egpgv19}
C.~Gueunet, P.~Fortin, J.~Jomier, and J.~Tierny.
\newblock {Task-based Augmented Reeb Graphs with Dynamic ST-Trees}.
\newblock In {\em EGPGV}, pp. 27--37. EG, Eindhoven, 2019.
  \href{https://doi.org/10.2312/PGV.20191107}
{doi: {{%
10\hspace{.1pt}\discretionary{.}{%
}{.}\hspace{.4pt}2312\discretionary{/}{%
}{/}PGV\hspace{.1pt}\discretionary{.}{%
}{.}\hspace{.4pt}20191107}}}


\bibitem{guillou_tvcg23}
P.~Guillou, J.~Vidal, and J.~Tierny.
\newblock {Discrete Morse Sandwich: Fast Computation of Persistence Diagrams
  for Scalar Data -- An Algorithm and A Benchmark}.
\newblock {\em IEEE TVCG}, 30(4):1897--1915, 2023.
  \href{https://doi.org/10.1109/TVCG.2023.3238008}
{doi: {{%
10\hspace{.1pt}\discretionary{.}{%
}{.}\hspace{.4pt}1109\discretionary{/}{%
}{/}TVCG\hspace{.1pt}\discretionary{.}{%
}{.}\hspace{.4pt}2023\hspace{.1pt}\discretionary{.}{%
}{.}\hspace{.4pt}3238008}}}


\bibitem{GuntherRSW14}
D.~G{\"{u}}nther, J.~Reininghaus, H.~Seidel, and T.~Weinkauf.
\newblock Notes on the simplification of the morse-smale complex.
\newblock In {\em TopoInVis}, pp. 135--150. Springer, Berlin, 2013.
  \href{https://doi.org/10.1007/978-3-319-04099-8_9}
{doi: {{%
10\hspace{.1pt}\discretionary{.}{%
}{.}\hspace{.4pt}1007\discretionary{/}{%
}{/}978\discretionary{%
}{-}{-}3\discretionary{%
}{-}{-}319\discretionary{%
}{-}{-}04099\discretionary{%
}{-}{-}8\_9}}}


\bibitem{mandatory}
D.~G{\"u}nther, J.~Salmon, and J.~Tierny.
\newblock Mandatory critical points of 2{D} uncertain scalar fields.
\newblock {\em CGF}, 33:31--40, 2014. \href{https://doi.org/10.1111/cgf.12359}
{doi: {{%
10\hspace{.1pt}\discretionary{.}{%
}{.}\hspace{.4pt}1111\discretionary{/}{%
}{/}cgf\hspace{.1pt}\discretionary{.}{%
}{.}\hspace{.4pt}12359}}}


\bibitem{gyulassy_ev14}
A.~Gyulassy, P.~Bremer, R.~Grout, H.~Kolla, J.~Chen, and V.~Pascucci.
\newblock Stability of dissipation elements: A case study in combustion.
\newblock {\em CGF}, 33(3):51--60, 2014.
  \href{https://doi.org/10.1111/CGF.12361}
{doi: {{%
10\hspace{.1pt}\discretionary{.}{%
}{.}\hspace{.4pt}1111\discretionary{/}{%
}{/}CGF\hspace{.1pt}\discretionary{.}{%
}{.}\hspace{.4pt}12361}}}


\bibitem{GyulassyBHP11}
A.~Gyulassy, P.~Bremer, B.~Hamann, and V.~Pascucci.
\newblock Practical considerations in morse-smale complex computation.
\newblock In {\em TopoInVis}, pp. 67--78. Springer, Berlin, 2009.
  \href{https://doi.org/10.1007/978-3-642-15014-2_6}
{doi: {{%
10\hspace{.1pt}\discretionary{.}{%
}{.}\hspace{.4pt}1007\discretionary{/}{%
}{/}978\discretionary{%
}{-}{-}3\discretionary{%
}{-}{-}642\discretionary{%
}{-}{-}15014\discretionary{%
}{-}{-}2\_6}}}


\bibitem{GyulassyBP12}
A.~Gyulassy, P.~Bremer, and V.~Pascucci.
\newblock Computing morse-smale complexes with accurate geometry.
\newblock {\em IEEE TVCG}, 18(12):2014--2022, 2012.
  \href{https://doi.org/10.1109/TVCG.2012.209}
{doi: {{%
10\hspace{.1pt}\discretionary{.}{%
}{.}\hspace{.4pt}1109\discretionary{/}{%
}{/}TVCG\hspace{.1pt}\discretionary{.}{%
}{.}\hspace{.4pt}2012\hspace{.1pt}\discretionary{.}{%
}{.}\hspace{.4pt}209}}}


\bibitem{gyulassy_vis18}
A.~Gyulassy, P.~T. Bremer, and V.~Pascucci.
\newblock Shared-memory parallel computation of morse-smale complexes with
  improved accuracy.
\newblock {\em IEEE TVCG}, 25(1):1183--1192, 2019.
  \href{https://doi.org/10.1109/TVCG.2018.2864848}
{doi: {{%
10\hspace{.1pt}\discretionary{.}{%
}{.}\hspace{.4pt}1109\discretionary{/}{%
}{/}TVCG\hspace{.1pt}\discretionary{.}{%
}{.}\hspace{.4pt}2018\hspace{.1pt}\discretionary{.}{%
}{.}\hspace{.4pt}2864848}}}


\bibitem{gyulassy_vis15}
A.~Gyulassy, A.~Knoll, K.~Lau, B.~Wang, P.~Bremer, M.~Papka, L.~A. Curtiss, and
  V.~Pascucci.
\newblock {Interstitial and Interlayer Ion Diffusion Geometry Extraction in
  Graphitic Nanosphere Battery Materials}.
\newblock {\em IEEE TVCG}, 22(1):916--925, 2016.
  \href{https://doi.org/10.1109/TVCG.2015.2467432}
{doi: {{%
10\hspace{.1pt}\discretionary{.}{%
}{.}\hspace{.4pt}1109\discretionary{/}{%
}{/}TVCG\hspace{.1pt}\discretionary{.}{%
}{.}\hspace{.4pt}2015\hspace{.1pt}\discretionary{.}{%
}{.}\hspace{.4pt}2467432}}}


\bibitem{chemistry_vis14}
D.~Günther, R.~A. Boto, J.~Contreras-Garcia, J.-P. Piquemal, and J.~Tierny.
\newblock Characterizing molecular interactions in chemical systems.
\newblock {\em IEEE TVCG}, 20(12):2476--2485, 2014.
  \href{https://doi.org/10.1109/TVCG.2014.2346403}
{doi: {{%
10\hspace{.1pt}\discretionary{.}{%
}{.}\hspace{.4pt}1109\discretionary{/}{%
}{/}TVCG\hspace{.1pt}\discretionary{.}{%
}{.}\hspace{.4pt}2014\hspace{.1pt}\discretionary{.}{%
}{.}\hspace{.4pt}2346403}}}


\bibitem{freudenthal42}
{H. Freudenthal}.
\newblock {Simplizialzerlegungen von beschrankter Flachheit}.
\newblock {\em {Ann. of Math.}}, {43}(3):580--583, 1942.
  \href{https://doi.org/10.48550/arXiv.2302.11922}
{doi: {{%
10\hspace{.1pt}\discretionary{.}{%
}{.}\hspace{.4pt}48550\discretionary{/}{%
}{/}arXiv\hspace{.1pt}\discretionary{.}{%
}{.}\hspace{.4pt}2302\hspace{.1pt}\discretionary{.}{%
}{.}\hspace{.4pt}11922}}}


\bibitem{hadad.etal_ijqc_2019}
C.~Hadad, E.~Florez, N.~Acelas, G.~Merino, and A.~Restrepo.
\newblock Microsolvation of small cations and anions.
\newblock {\em IJQC}, 119(2):e25766, 2019.
  \href{https://doi.org/10.1002/qua.25766}
{doi: {{%
10\hspace{.1pt}\discretionary{.}{%
}{.}\hspace{.4pt}1002\discretionary{/}{%
}{/}qua\hspace{.1pt}\discretionary{.}{%
}{.}\hspace{.4pt}25766}}}


\bibitem{avogadro}
M.~D. Hanwell, D.~E. Curtis, D.~C. Lonie, T.~Vandermeersch, E.~Zurek, and G.~R.
  Hutchison.
\newblock Avogadro: an advanced semantic chemical editor, visualization, and
  analysis platform.
\newblock {\em J. of Cheminfo.}, 4(1):17, 2012.
  \href{https://doi.org/10.1186/1758-2946-4-17}
{doi: {{%
10\hspace{.1pt}\discretionary{.}{%
}{.}\hspace{.4pt}1186\discretionary{/}{%
}{/}1758\discretionary{%
}{-}{-}2946\discretionary{%
}{-}{-}4\discretionary{%
}{-}{-}17}}}


\bibitem{heine16}
C.~Heine, H.~Leitte, M.~Hlawitschka, F.~Iuricich, L.~De~Floriani,
  G.~Scheuermann, H.~Hagen, and C.~Garth.
\newblock A survey of topology-based methods in visualization.
\newblock {\em CGF}, 35(3):643--667, 2016.
  \href{https://doi.org/10.1111/CGF.12933}
{doi: {{%
10\hspace{.1pt}\discretionary{.}{%
}{.}\hspace{.4pt}1111\discretionary{/}{%
}{/}CGF\hspace{.1pt}\discretionary{.}{%
}{.}\hspace{.4pt}12933}}}


\bibitem{hohenberg64}
P.~Hohenberg and W.~Kohn.
\newblock Inhomogeneous electron gas.
\newblock {\em Phys. Rev.}, 136:B864--B871, 1964.
  \href{https://doi.org/10.1103/PhysRev.136.B864}
{doi: {{%
10\hspace{.1pt}\discretionary{.}{%
}{.}\hspace{.4pt}1103\discretionary{/}{%
}{/}PhysRev\hspace{.1pt}\discretionary{.}{%
}{.}\hspace{.4pt}136\hspace{.1pt}\discretionary{.}{%
}{.}\hspace{.4pt}B864}}}


\bibitem{kasten_tvcg11}
J.~Kasten, J.~Reininghaus, I.~Hotz, and H.~Hege.
\newblock Two-dimensional time-dependent vortex regions based on the
  acceleration magnitude.
\newblock {\em IEEE TVCG}, 17(12):2080--2087, 2011.
  \href{https://doi.org/10.1109/TVCG.2011.249}
{doi: {{%
10\hspace{.1pt}\discretionary{.}{%
}{.}\hspace{.4pt}1109\discretionary{/}{%
}{/}TVCG\hspace{.1pt}\discretionary{.}{%
}{.}\hspace{.4pt}2011\hspace{.1pt}\discretionary{.}{%
}{.}\hspace{.4pt}249}}}


\bibitem{KozlikovaKFLBBV17}
B.~Kozl{\'{\i}}kov{\'{a}}, M.~Krone, M.~Falk, N.~Lindow, M.~Baaden, D.~Baum,
  I.~Viola, J.~Parulek, and H.~Hege.
\newblock Visualization of biomolecular structures: State of the art revisited.
\newblock {\em CGF}, 36(8):178--204, 2017.
  \href{https://doi.org/10.1111/cgf.13072}
{doi: {{%
10\hspace{.1pt}\discretionary{.}{%
}{.}\hspace{.4pt}1111\discretionary{/}{%
}{/}cgf\hspace{.1pt}\discretionary{.}{%
}{.}\hspace{.4pt}13072}}}


\bibitem{chemistry33}
M.~Krone, M.~Falk, S.~Rehm, J.~Pleiss, and T.~Ertl.
\newblock {Interactive Exploration of Protein Cavities}.
\newblock {\em CGF}, 30(3):673--682, 2011.
  \href{https://doi.org/10.1111/j.1467-8659.2011.01916.x}
{doi: {{%
10\hspace{.1pt}\discretionary{.}{%
}{.}\hspace{.4pt}1111\discretionary{/}{%
}{/}j\hspace{.1pt}\discretionary{.}{%
}{.}\hspace{.4pt}1467\discretionary{%
}{-}{-}8659\hspace{.1pt}\discretionary{.}{%
}{.}\hspace{.4pt}2011\hspace{.1pt}\discretionary{.}{%
}{.}\hspace{.4pt}01916\hspace{.1pt}\discretionary{.}{%
}{.}\hspace{.4pt}x}}}


\bibitem{chemistry34}
M.~Krone, G.~Reina, C.~Schulz, T.~Kulschewski, J.~Pleiss, and T.~Ertl.
\newblock Interactive extraction and tracking of biomolecular surface features.
\newblock {\em CGF}, 32(3pt3):331--340, 2013.
  \href{https://doi.org/10.1111/cgf.12120}
{doi: {{%
10\hspace{.1pt}\discretionary{.}{%
}{.}\hspace{.4pt}1111\discretionary{/}{%
}{/}cgf\hspace{.1pt}\discretionary{.}{%
}{.}\hspace{.4pt}12120}}}


\bibitem{chemistry37}
N.~Lindow, D.~Baum, and H.-C. Hege.
\newblock Voronoi-based extraction and visualization of molecular paths.
\newblock {\em IEEE TVCG}, 17(12):2025--2034, 2011.
  \href{https://doi.org/10.1109/TVCG.2011.259}
{doi: {{%
10\hspace{.1pt}\discretionary{.}{%
}{.}\hspace{.4pt}1109\discretionary{/}{%
}{/}TVCG\hspace{.1pt}\discretionary{.}{%
}{.}\hspace{.4pt}2011\hspace{.1pt}\discretionary{.}{%
}{.}\hspace{.4pt}259}}}


\bibitem{Lukasczyk_vis20}
J.~Lukasczyk, C.~Garth, R.~Maciejewski, and J.~Tierny.
\newblock Localized topological simplification of scalar data.
\newblock {\em IEEE TVCG}, 27(2):572--582, 2020.
  \href{https://doi.org/10.1109/TVCG.2020.3030353}
{doi: {{%
10\hspace{.1pt}\discretionary{.}{%
}{.}\hspace{.4pt}1109\discretionary{/}{%
}{/}TVCG\hspace{.1pt}\discretionary{.}{%
}{.}\hspace{.4pt}2020\hspace{.1pt}\discretionary{.}{%
}{.}\hspace{.4pt}3030353}}}


\bibitem{MasoodTLANH21}
T.~B. Masood, S.~Thygesen, M.~Linares, A.~I. Abrikosov, V.~Natarajan, and
  I.~Hotz.
\newblock Visual analysis of electronic densities and transitions in molecules.
\newblock {\em CGF}, 40(3):287--298, 2021.
  \href{https://doi.org/10.1111/cgf.14307}
{doi: {{%
10\hspace{.1pt}\discretionary{.}{%
}{.}\hspace{.4pt}1111\discretionary{/}{%
}{/}cgf\hspace{.1pt}\discretionary{.}{%
}{.}\hspace{.4pt}14307}}}


\bibitem{matta07}
C.~{Matta} and R.~{Boyd}.
\newblock {\em The Quantum Theory of Atoms in Molecules: From Solid State to
  DNA and Drug Design}.
\newblock Wiley-VCH, 2007.

\bibitem{milnor63}
J.~Milnor.
\newblock {\em {Morse Theory}}.
\newblock 1963.

\bibitem{Munkres1957}
J.~Munkres.
\newblock Algorithms for the assignment and transportation problems.
\newblock {\em J. of SIAM}, 5(1):32--38, 1957.
  \href{https://doi.org/10.1137/0105003}
{doi: {{%
10\hspace{.1pt}\discretionary{.}{%
}{.}\hspace{.4pt}1137\discretionary{/}{%
}{/}0105003}}}


\bibitem{NarayananTN15}
V.~Narayanan, D.~M. Thomas, and V.~Natarajan.
\newblock Distance between extremum graphs.
\newblock In {\em PacificVis}, pp. 263--270, 2015.
  \href{https://doi.org/10.1109/PACIFICVIS.2015.7156386}
{doi: {{%
10\hspace{.1pt}\discretionary{.}{%
}{.}\hspace{.4pt}1109\discretionary{/}{%
}{/}PACIFICVIS\hspace{.1pt}\discretionary{.}{%
}{.}\hspace{.4pt}2015\hspace{.1pt}\discretionary{.}{%
}{.}\hspace{.4pt}7156386}}}


\bibitem{chemistry42}
V.~Natarajan, Y.~Wang, P.-T. Bremer, V.~Pascucci, and B.~Hamann.
\newblock Segmenting molecular surfaces.
\newblock {\em CAGD}, 23(6):495--509, 2006.
  \href{https://doi.org/10.1016/j.cagd.2006.02.003}
{doi: {{%
10\hspace{.1pt}\discretionary{.}{%
}{.}\hspace{.4pt}1016\discretionary{/}{%
}{/}j\hspace{.1pt}\discretionary{.}{%
}{.}\hspace{.4pt}cagd\hspace{.1pt}\discretionary{.}{%
}{.}\hspace{.4pt}2006\hspace{.1pt}\discretionary{.}{%
}{.}\hspace{.4pt}02\hspace{.1pt}\discretionary{.}{%
}{.}\hspace{.4pt}003}}}


\bibitem{nauleau_ldav22}
F.~Nauleau, F.~Vivodtzev, T.~Bridel-Bertomeu, H.~Beaugendre, and J.~Tierny.
\newblock {Topological Analysis of Ensembles of Hydrodynamic Turbulent Flows --
  An Experimental Study}.
\newblock In {\em IEEE LDAV}, pp. 1--11. Los Alamitos, 2022.
  \href{https://doi.org/10.1109/LDAV57265.2022.9966403}
{doi: {{%
10\hspace{.1pt}\discretionary{.}{%
}{.}\hspace{.4pt}1109\discretionary{/}{%
}{/}LDAV57265\hspace{.1pt}\discretionary{.}{%
}{.}\hspace{.4pt}2022\hspace{.1pt}\discretionary{.}{%
}{.}\hspace{.4pt}9966403}}}


\bibitem{nauta.miller_s_2000}
K.~Nauta and R.~E. Miller.
\newblock Formation of {{Cyclic Water Hexamer}} in {{Liquid Helium}}: {{The
  Smallest Piece}} of {{Ice}}.
\newblock {\em Science}, 287(5451):293--295, 2000.
  \href{https://doi.org/10.1126/science.287.5451.293}
{doi: {{%
10\hspace{.1pt}\discretionary{.}{%
}{.}\hspace{.4pt}1126\discretionary{/}{%
}{/}science\hspace{.1pt}\discretionary{.}{%
}{.}\hspace{.4pt}287\hspace{.1pt}\discretionary{.}{%
}{.}\hspace{.4pt}5451\hspace{.1pt}\discretionary{.}{%
}{.}\hspace{.4pt}293}}}


\bibitem{Malgorzata19}
M.~Olejniczak, A.~S.~P. Gomes, and J.~Tierny.
\newblock {A Topological Data Analysis Perspective on Non-Covalent Interactions
  in Relativistic Calculations}.
\newblock {\em IJQC}, 120(8):e26133, 2019.
  \href{https://doi.org/10.1002/qua.26133}
{doi: {{%
10\hspace{.1pt}\discretionary{.}{%
}{.}\hspace{.4pt}1002\discretionary{/}{%
}{/}qua\hspace{.1pt}\discretionary{.}{%
}{.}\hspace{.4pt}26133}}}


\bibitem{chemistry45}
J.~Parulek and A.~Brambilla.
\newblock Fast blending scheme for molecular surface representation.
\newblock {\em IEEE TVCG}, 19(12):2653--2662, 2013.
  \href{https://doi.org/10.1109/TVCG.2013.158}
{doi: {{%
10\hspace{.1pt}\discretionary{.}{%
}{.}\hspace{.4pt}1109\discretionary{/}{%
}{/}TVCG\hspace{.1pt}\discretionary{.}{%
}{.}\hspace{.4pt}2013\hspace{.1pt}\discretionary{.}{%
}{.}\hspace{.4pt}158}}}


\bibitem{chemistry46}
J.~Parulek, C.~Turkay, N.~Reuter, and I.~Viola.
\newblock Implicit surfaces for interactive graph based cavity analysis of
  molecular simulations.
\newblock In {\em BioVis}, pp. 115--122, 2012.
  \href{https://doi.org/10.1109/BioVis.2012.6378601}
{doi: {{%
10\hspace{.1pt}\discretionary{.}{%
}{.}\hspace{.4pt}1109\discretionary{/}{%
}{/}BioVis\hspace{.1pt}\discretionary{.}{%
}{.}\hspace{.4pt}2012\hspace{.1pt}\discretionary{.}{%
}{.}\hspace{.4pt}6378601}}}


\bibitem{PeyreC19}
G.~Peyr{\'{e}} and M.~Cuturi.
\newblock Computational optimal transport.
\newblock {\em Found. Trends Mach. Learn.}, 2019.

\bibitem{perez.etal_s_2012}
C.~Pérez, M.~T. Muckle, D.~P. Zaleski, and et~al.
\newblock Structures of cage, prism, and book isomers of water hexamer from
  broadband rotational spectroscopy.
\newblock {\em Science}, 336(6083):897--901, 2012.
  \href{https://doi.org/10.1126/science.1220574}
{doi: {{%
10\hspace{.1pt}\discretionary{.}{%
}{.}\hspace{.4pt}1126\discretionary{/}{%
}{/}science\hspace{.1pt}\discretionary{.}{%
}{.}\hspace{.4pt}1220574}}}


\bibitem{quack.seyfang_msaqd_2021}
M.~Quack and G.~Seyfang.
\newblock Chapter 7 - {Atomic} and {Molecular} {Tunneling} {Processes} in
  {Chemistry}.
\newblock In R.~Marquardt and M.~Quack, eds., {\em Molecular {Spectroscopy} and
  {Quantum} {Dynamics}}. 2021.

\bibitem{empiricallyConvergent}
J.~Reininghaus, D.~G{\"{u}}nther, I.~Hotz, T.~Weinkauf, and H.~Seidel.
\newblock Combinatorial gradient fields for 2d images with empirically
  convergent separatrices.
\newblock {\em CoRR}, abs/1208.6523, 2012.

\bibitem{richardson_tjocp_2018}
J.~O. Richardson.
\newblock Perspective: {Ring}-polymer instanton theory.
\newblock {\em JCP}, 148(20):200901, 2018.
  \href{https://doi.org/10.1063/1.5028352}
{doi: {{%
10\hspace{.1pt}\discretionary{.}{%
}{.}\hspace{.4pt}1063\discretionary{/}{%
}{/}1\hspace{.1pt}\discretionary{.}{%
}{.}\hspace{.4pt}5028352}}}


\bibitem{rotations16}
J.~O. Richardson, C.~Pérez, S.~Lobsiger, A.~A. Reid, B.~Temelso, G.~C.
  Shields, Z.~Kisiel, D.~J. Wales, B.~H. Pate, and S.~C. Althorpe.
\newblock Concerted hydrogen-bond breaking by quantum tunneling in the water
  hexamer prism.
\newblock {\em Science}, 351:1310--1313, 2016.
  \href{https://doi.org/10.1126/science.aae0012}
{doi: {{%
10\hspace{.1pt}\discretionary{.}{%
}{.}\hspace{.4pt}1126\discretionary{/}{%
}{/}science\hspace{.1pt}\discretionary{.}{%
}{.}\hspace{.4pt}aae0012}}}


\bibitem{robins_pami11}
V.~Robins, P.~J. Wood, and A.~P. Sheppard.
\newblock {Theory and Algorithms for Constructing Discrete Morse Complexes from
  Grayscale Digital Images}.
\newblock {\em {IEEE} PAMI}, 33:1646--1658, 2011.
  \href{https://doi.org/10.1109/TPAMI.2011.95}
{doi: {{%
10\hspace{.1pt}\discretionary{.}{%
}{.}\hspace{.4pt}1109\discretionary{/}{%
}{/}TPAMI\hspace{.1pt}\discretionary{.}{%
}{.}\hspace{.4pt}2011\hspace{.1pt}\discretionary{.}{%
}{.}\hspace{.4pt}95}}}


\bibitem{santis.etal_tjocp_2024}
G.~D. Santis, K.~M. Herman, J.~P. Heindel, and S.~S. Xantheas.
\newblock Descriptors of water aggregation.
\newblock {\em JCP}, 160:054306, 2024. \href{https://doi.org/10.1063/5.0179815}
{doi: {{%
10\hspace{.1pt}\discretionary{.}{%
}{.}\hspace{.4pt}1063\discretionary{/}{%
}{/}5\hspace{.1pt}\discretionary{.}{%
}{.}\hspace{.4pt}0179815}}}


\bibitem{saykally.wales_s_2012}
R.~J. Saykally and D.~J. Wales.
\newblock Pinning {Down} the {Water} {Hexamer}.
\newblock {\em Science}, 336(6083):814--815, 2012.
  \href{https://doi.org/10.1126/science.1222007}
{doi: {{%
10\hspace{.1pt}\discretionary{.}{%
}{.}\hspace{.4pt}1126\discretionary{/}{%
}{/}science\hspace{.1pt}\discretionary{.}{%
}{.}\hspace{.4pt}1222007}}}


\bibitem{SharmaMTLHN24}
M.~Sharma, T.~B. Masood, S.~S. Thygesen, M.~Linares, I.~Hotz, and V.~Natarajan.
\newblock Continuous scatterplot operators for bivariate analysis and study of
  electronic transitions.
\newblock {\em IEEE TVCG}, 30(7):3532--3544, 2024.
  \href{https://doi.org/10.1109/TVCG.2023.3237768}
{doi: {{%
10\hspace{.1pt}\discretionary{.}{%
}{.}\hspace{.4pt}1109\discretionary{/}{%
}{/}TVCG\hspace{.1pt}\discretionary{.}{%
}{.}\hspace{.4pt}2023\hspace{.1pt}\discretionary{.}{%
}{.}\hspace{.4pt}3237768}}}


\bibitem{ShivashankarN12}
N.~Shivashankar and V.~Natarajan.
\newblock {Parallel Computation of 3D Morse-Smale Complexes}.
\newblock {\em CGF}, 31(3):965--974, 2012.
  \href{https://doi.org/10.1111/J.1467-8659.2012.03089.X}
{doi: {{%
10\hspace{.1pt}\discretionary{.}{%
}{.}\hspace{.4pt}1111\discretionary{/}{%
}{/}J\hspace{.1pt}\discretionary{.}{%
}{.}\hspace{.4pt}1467\discretionary{%
}{-}{-}8659\hspace{.1pt}\discretionary{.}{%
}{.}\hspace{.4pt}2012\hspace{.1pt}\discretionary{.}{%
}{.}\hspace{.4pt}03089\hspace{.1pt}\discretionary{.}{%
}{.}\hspace{.4pt}X}}}


\bibitem{shivashankar2016felix}
N.~Shivashankar, P.~Pranav, V.~Natarajan, R.~van~de Weygaert, E.~P. Bos, and
  S.~Rieder.
\newblock Felix: A topology based framework for visual exploration of cosmic
  filaments.
\newblock {\em IEEE TVCG}, 22(6):1745--1759, 2016.
  \href{https://doi.org/10.1109/TVCG.2015.2452919}
{doi: {{%
10\hspace{.1pt}\discretionary{.}{%
}{.}\hspace{.4pt}1109\discretionary{/}{%
}{/}TVCG\hspace{.1pt}\discretionary{.}{%
}{.}\hspace{.4pt}2015\hspace{.1pt}\discretionary{.}{%
}{.}\hspace{.4pt}2452919}}}


\bibitem{SkanbergFLYH22}
R.~Sk{\aa}nberg, M.~Falk, M.~Linares, A.~Ynnerman, and I.~Hotz.
\newblock Tracking internal frames of reference for consistent molecular
  distribution functions.
\newblock {\em IEEE TVCG}, 28(9):3126--3137, 2022.
  \href{https://doi.org/10.1109/TVCG.2021.3051632}
{doi: {{%
10\hspace{.1pt}\discretionary{.}{%
}{.}\hspace{.4pt}1109\discretionary{/}{%
}{/}TVCG\hspace{.1pt}\discretionary{.}{%
}{.}\hspace{.4pt}2021\hspace{.1pt}\discretionary{.}{%
}{.}\hspace{.4pt}3051632}}}


\bibitem{SkanbergHYL23}
R.~Sk{\aa}nberg, I.~Hotz, A.~Ynnerman, and M.~Linares.
\newblock {VIAMD:} a software for visual interactive analysis of molecular
  dynamics.
\newblock {\em J. Chem. Inf. Model.}, 63(23):7382--7391, 2023.
  \href{https://doi.org/10.1021/acs.jcim.3c01033}
{doi: {{%
10\hspace{.1pt}\discretionary{.}{%
}{.}\hspace{.4pt}1021\discretionary{/}{%
}{/}acs\hspace{.1pt}\discretionary{.}{%
}{.}\hspace{.4pt}jcim\hspace{.1pt}\discretionary{.}{%
}{.}\hspace{.4pt}3c01033}}}


\bibitem{SkanbergLFHY19}
R.~Sk{\aa}nberg, M.~Linares, M.~Falk, I.~Hotz, and A.~Ynnerman.
\newblock Molfind - integrated multi-selection schemes for complex molecular
  structures.
\newblock In {\em MolVA}. Eurographics, 2019.
  \href{https://doi.org/10.2312/molva.20191096}
{doi: {{%
10\hspace{.1pt}\discretionary{.}{%
}{.}\hspace{.4pt}2312\discretionary{/}{%
}{/}molva\hspace{.1pt}\discretionary{.}{%
}{.}\hspace{.4pt}20191096}}}


\bibitem{soler_ldav19}
M.~Soler, M.~Petitfrere, G.~Darche, M.~Plainchault, B.~Conche, and J.~Tierny.
\newblock {Ranking Viscous Finger Simulations to an Acquired Ground Truth with
  Topology-Aware Matchings}.
\newblock In {\em IEEE LDAV}, pp. 62--72. IEEE, Los Alamitos, 2019.
  \href{https://doi.org/10.1109/LDAV48142.2019.8944365}
{doi: {{%
10\hspace{.1pt}\discretionary{.}{%
}{.}\hspace{.4pt}1109\discretionary{/}{%
}{/}LDAV48142\hspace{.1pt}\discretionary{.}{%
}{.}\hspace{.4pt}2019\hspace{.1pt}\discretionary{.}{%
}{.}\hspace{.4pt}8944365}}}


\bibitem{taylor_cg&d_2024}
R.~Taylor.
\newblock Aerogen {Bond}, {Halogen} {Bond}, {Chalcogen} {Bond}, {Pnictogen}
  {Bond}, {Tetrel} {Bond}, {Triel} {Bond} ... {Why} {So} {Many} {Names}?
\newblock {\em Crystal Growth \& Design}, 24(10):4003--4012, 2024.
  \href{https://doi.org/10.1021/acs.cgd.4c00303}
{doi: {{%
10\hspace{.1pt}\discretionary{.}{%
}{.}\hspace{.4pt}1021\discretionary{/}{%
}{/}acs\hspace{.1pt}\discretionary{.}{%
}{.}\hspace{.4pt}cgd\hspace{.1pt}\discretionary{.}{%
}{.}\hspace{.4pt}4c00303}}}


\bibitem{tevelde.etal_jcc_2001}
G.~Te~Velde, F.~M. Bickelhaupt, E.~J. Baerends, C.~Fonseca~Guerra, S.~J.~A.
  Van~Gisbergen, J.~G. Snijders, and T.~Ziegler.
\newblock Chemistry with {{ADF}}.
\newblock {\em J. of Comp. Chem.}, 22(9):931--967, 2001.
  \href{https://doi.org/10.1002/jcc.1056}
{doi: {{%
10\hspace{.1pt}\discretionary{.}{%
}{.}\hspace{.4pt}1002\discretionary{/}{%
}{/}jcc\hspace{.1pt}\discretionary{.}{%
}{.}\hspace{.4pt}1056}}}


\bibitem{ThanhAW24}
S.~L. Thanh, M.~Ankele, and T.~Weinkauf.
\newblock Revisiting accurate geometry for morse-smale complexes.
\newblock In {\em {IEEE} TopoInVis}, pp. 34--43. IEEE, Los Alamitos, 2024.
  \href{https://doi.org/10.1109/TopoInVis64104.2024.00008}
{doi: {{%
10\hspace{.1pt}\discretionary{.}{%
}{.}\hspace{.4pt}1109\discretionary{/}{%
}{/}TopoInVis64104\hspace{.1pt}\discretionary{.}{%
}{.}\hspace{.4pt}2024\hspace{.1pt}\discretionary{.}{%
}{.}\hspace{.4pt}00008}}}


\bibitem{ttk17}
J.~Tierny, G.~Favelier, J.~A. Levine, C.~Gueunet, and M.~Michaux.
\newblock {The Topology ToolKit}.
\newblock {\em IEEE TVCG}, 24(1):832--842, 2017.
  \href{https://doi.org/10.1109/TVCG.2017.2743938}
{doi: {{%
10\hspace{.1pt}\discretionary{.}{%
}{.}\hspace{.4pt}1109\discretionary{/}{%
}{/}TVCG\hspace{.1pt}\discretionary{.}{%
}{.}\hspace{.4pt}2017\hspace{.1pt}\discretionary{.}{%
}{.}\hspace{.4pt}2743938}}}


\bibitem{chemistry54}
M.~van~der Zwan, W.~Lueks, H.~Bekker, and T.~Isenberg.
\newblock Illustrative molecular visualiszation with continuous abstraction.
\newblock {\em CGF}, 30(3):683--690, 2011.
  \href{https://doi.org/10.1111/j.1467-8659.2011.01917.x}
{doi: {{%
10\hspace{.1pt}\discretionary{.}{%
}{.}\hspace{.4pt}1111\discretionary{/}{%
}{/}j\hspace{.1pt}\discretionary{.}{%
}{.}\hspace{.4pt}1467\discretionary{%
}{-}{-}8659\hspace{.1pt}\discretionary{.}{%
}{.}\hspace{.4pt}2011\hspace{.1pt}\discretionary{.}{%
}{.}\hspace{.4pt}01917\hspace{.1pt}\discretionary{.}{%
}{.}\hspace{.4pt}x}}}


\bibitem{wang.etal_nr_2016}
B.~Wang, W.~Jiang, Y.~Gao, B.~K. Teo, and Z.~Wang.
\newblock Chirality recognition in concerted proton transfer process for
  prismatic water clusters.
\newblock {\em Nano Research}, 9(9):2782--2795, 2016.
  \href{https://doi.org/10.1007/s12274-016-1167-x}
{doi: {{%
10\hspace{.1pt}\discretionary{.}{%
}{.}\hspace{.4pt}1007\discretionary{/}{%
}{/}s12274\discretionary{%
}{-}{-}016\discretionary{%
}{-}{-}1167\discretionary{%
}{-}{-}x}}}


\bibitem{wilson_tjocp_1939}
E.~B. Wilson, Jr.
\newblock A {{Method}} of {{Obtaining}} the {{Expanded Secular Equation}} for
  the {{Vibration Frequencies}} of a {{Molecule}}.
\newblock {\em The Journal of Chemical Physics}, 7(11):1047--1052, 1939.
  \href{https://doi.org/10.1063/1.1750363}
{doi: {{%
10\hspace{.1pt}\discretionary{.}{%
}{.}\hspace{.4pt}1063\discretionary{/}{%
}{/}1\hspace{.1pt}\discretionary{.}{%
}{.}\hspace{.4pt}1750363}}}


\bibitem{wolke.etal_s_2016}
C.~T. Wolke, J.~A. Fournier, L.~C. Dzugan, M.~R. Fagiani, T.~T. Odbadrakh,
  H.~Knorke, K.~D. Jordan, A.~B. McCoy, K.~R. Asmis, and M.~A. Johnson.
\newblock Spectroscopic snapshots of the proton-transfer mechanism in water.
\newblock {\em Science}, 354(6316):1131--1135, 2016.
  \href{https://doi.org/10.1126/science.aaf8425}
{doi: {{%
10\hspace{.1pt}\discretionary{.}{%
}{.}\hspace{.4pt}1126\discretionary{/}{%
}{/}science\hspace{.1pt}\discretionary{.}{%
}{.}\hspace{.4pt}aaf8425}}}


\bibitem{yang.etal_n_2021}
J.~Yang, R.~Dettori, J.~P.~F. Nunes, N.~H. List, E.~Biasin, M.~Centurion,
  Z.~Chen, A.~A. Cordones, D.~P. Deponte, T.~F. Heinz, M.~E. Kozina,
  K.~Ledbetter, M.-F. Lin, A.~M. Lindenberg, M.~Mo, A.~Nilsson, X.~Shen,
  T.~J.~A. Wolf, D.~Donadio, K.~J. Gaffney, T.~J. Martinez, and X.~Wang.
\newblock Direct observation of ultrafast hydrogen bond strengthening in liquid
  water.
\newblock {\em Nature}, 596:531--535, 2021.
  \href{https://doi.org/10.1038/s41586-021-03793-9}
{doi: {{%
10\hspace{.1pt}\discretionary{.}{%
}{.}\hspace{.4pt}1038\discretionary{/}{%
}{/}s41586\discretionary{%
}{-}{-}021\discretionary{%
}{-}{-}03793\discretionary{%
}{-}{-}9}}}


\bibitem{yoo.xantheas_hocc_2017}
S.~Yoo and S.~S. Xantheas.
\newblock Structures, {Energetics}, and {Spectroscopic} {Fingerprints} of
  {Water} {Clusters} n = 2–24.
\newblock In {\em Handbook of {Computational} {Chemistry}}. 2017.
  \href{https://doi.org/10.1007/978-3-319-27282-5_21}
{doi: {{%
10\hspace{.1pt}\discretionary{.}{%
}{.}\hspace{.4pt}1007\discretionary{/}{%
}{/}978\discretionary{%
}{-}{-}3\discretionary{%
}{-}{-}319\discretionary{%
}{-}{-}27282\discretionary{%
}{-}{-}5\_21}}}


\bibitem{adamo.barone_tjocp_1999}
C.~Zdamo and V.~Barone.
\newblock Toward reliable density functional methods without adjustable
  parameters: {{The PBE0}} model.
\newblock {\em The Journal of Chemical Physics}, 110(13):6158--6170, 1999.
  \href{https://doi.org/10.1063/1.478522}
{doi: {{%
10\hspace{.1pt}\discretionary{.}{%
}{.}\hspace{.4pt}1063\discretionary{/}{%
}{/}1\hspace{.1pt}\discretionary{.}{%
}{.}\hspace{.4pt}478522}}}


\bibitem{zhang.xu_a_2022}
C.~Zhang and W.~Xu.
\newblock Interactions between water and organic molecules or inorganic salts
  on surfaces.
\newblock {\em Aggregate}, 3(4):e175, 2022.
  \href{https://doi.org/10.1002/agt2.175}
{doi: {{%
10\hspace{.1pt}\discretionary{.}{%
}{.}\hspace{.4pt}1002\discretionary{/}{%
}{/}agt2\hspace{.1pt}\discretionary{.}{%
}{.}\hspace{.4pt}175}}}


\bibitem{chemistry58}
X.~Zhang and C.~Bajaj.
\newblock Extraction, quantification and visualization of protein pockets.
\newblock In {\em IEEE CSBC}, 2007.

\bibitem{zhao.etal_nrc_2019a}
L.~Zhao, W.~H.~E. Schwarz, and G.~Frenking.
\newblock The {{Lewis}} electron-pair bonding model: The physical background,
  one century later.
\newblock {\em Nature Reviews Chemistry}, 3:35--47, 2019.
  \href{https://doi.org/10.1038/s41570-018-0052-4}
{doi: {{%
10\hspace{.1pt}\discretionary{.}{%
}{.}\hspace{.4pt}1038\discretionary{/}{%
}{/}s41570\discretionary{%
}{-}{-}018\discretionary{%
}{-}{-}0052\discretionary{%
}{-}{-}4}}}


\bibitem{zomorodianBook}
A.~J. Zomorodian.
\newblock Topology for computing.
\newblock In {\em Algorithms and Theory of Computation Handbook (Second
  Edition)}. 2010.

\end{thebibliography}


\end{document}